%% file: main.tex
\documentclass{article}

    \PassOptionsToPackage{numbers, compress}{natbib}

\usepackage[preprint]{neurips_2026}

\usepackage[utf8]{inputenc} 
\usepackage[T1]{fontenc}    
\usepackage{hyperref}       
\usepackage{url}            
\usepackage{xurl}
\usepackage{ragged2e}
\usepackage{booktabs}       
\usepackage{amsfonts}       
\usepackage{amsmath}        
\usepackage{amssymb}
\usepackage{nicefrac}       
\usepackage{microtype}      
\usepackage{xcolor}         
\usepackage{graphicx}       
\usepackage{multirow}       
\usepackage{bbm}
\usepackage{adjustbox}
\usepackage{tabularx}
\usepackage{array}
\usepackage[table]{xcolor}
\usepackage{subcaption}
\usepackage{tcolorbox}
\definecolor{ac-maroon}{HTML}{A61E4D}
\definecolor{darkblue}{rgb}{0.0, 0.0, 0.55}
\usepackage{placeins}

\usepackage{makecell}
\usepackage{enumitem}
\usepackage{textcomp}

\newcommand{\benchmark}{\textsc{Reflect}}

\tcbset{
  inputbox/.style={
    colback=blue!8, colframe=blue!35,
    boxrule=0.5pt, arc=3pt,
    fonttitle=\bfseries\small, title={User Input},
    left=4pt, right=4pt, top=3pt, bottom=3pt
  },
  outputbox/.style={
    colback=orange!8, colframe=orange!40,
    boxrule=0.5pt, arc=3pt,
    fonttitle=\bfseries\small,
    left=4pt, right=4pt, top=3pt, bottom=3pt
  },
  promptbox/.style={
    colback=purple!6, colframe=purple!45,
    boxrule=0.5pt, arc=3pt,
    fonttitle=\bfseries\small,
    left=4pt, right=4pt, top=3pt, bottom=3pt
  }
}

\title{Time to \benchmark~: Can We Trust \\ LLM Judges for  Evidence-based Research Agents?}

\newcommand{\heart}{\ensuremath{\heartsuit}}

\author{
\begin{tabular}{cccc}
Leyao Wang$^{1,\heart,\dagger}$ &
Yanan He$^{1,\heart,\dagger}$ &
Peng Chen$^{1,\dagger}$ &
Asaf Yehudai$^{2,\dagger}$ \\
Yixin Liu$^{1}$ &
Rex Ying$^{1}$ &
Michal Shmueli-Scheuer$^{2}$ &
Arman Cohan$^{1,\dagger}$
\end{tabular}
\\[1.2em]
$^{1}$Yale University \qquad
$^{2}$IBM Research
\\[0.5em]
{\small \texttt{\{leyao.wang.lw855, yanan.he, peng.chen.pc838, yixin.liu, rex.ying, arman.cohan\}@yale.edu}} \\
{\small \texttt{Asaf.Yehudai@ibm.com, shmueli@il.ibm.com}}
\\[0.5em]
{\small $\heart$ Joint first authors. \quad $\dagger$ Core contributors. 
}
}

\begin{document}

\maketitle

\begingroup
\renewcommand\thefootnote{}
\footnotetext{See full author contributions \hyperref[sec:author]{here}.}
\endgroup

\input{sections/abstract}

\input{sections/intro}

\input{sections/method}

\input{sections/experiment}

\input{sections/related_works}
\input{sections/conclusion}

\input{sections/contribution}

\bibliographystyle{plainnat}
\bibliography{refs}
\newpage

\input{appendix/appendix}


\end{document}

%% file: sections/abstract.tex
 \begin{abstract}

Deep research agents increasingly automate complex information-seeking tasks, producing evidence-grounded reports via multi-step reasoning, tool use, and synthesis. Their growing role demands scalable, reliable evaluation, positioning LLM-as-judge as a supervision paradigm for assessing factual accuracy, evidence use, and reasoning quality. Yet the reliability of these judges for deep research agents remains poorly understood, posing a critical meta-evaluation problem: before deploying LLM judges to supervise research agents, we must first evaluate the judges themselves. Existing meta-evaluations fall short in two ways: (1) reliance on coarse, subjective human-preference agreement; (2) focus on instruction-following or verifiable tasks, leaving open-ended agent executions unexplored.
To address these gaps, we introduce \textbf{\benchmark{}}~(\textbf{RE}liable \textbf{F}ine-grained \textbf{L}LM judge \textbf{E}valuation via \textbf{C}ontrolled in\textbf{T}ervention), a meta-evaluation benchmark targeting fine-grained failure detection in agentic environments. \benchmark{} defines a detailed taxonomy of process- and outcome-level failure modes, instantiated by performing controlled and localized interventions on quality-screened agent execution traces. This yields \textit{verifiable}, \textit{comprehensive}, and \textit{fine-grained} instances for validating the judge models.
Our experiments show that current LLM judges remain unreliable: even the best-performing models achieve overall accuracies below 55\% across reasoning, tool-use, and report-quality failures, with especially poor performance on evidence verification. Together, our taxonomy and findings expose systematic judge limitations, reveal tradeoffs in cost and reliability, and offer actionable guidance for building more reliable evaluation pipelines for deep research agents.

\end{abstract}

%% file: sections/intro.tex
\section{Introduction}
\label{sec:intro}

Deep research agents are increasingly important for automating complex information-seeking tasks. They can investigate open-ended questions through browser interaction, reasoning, and synthesis, ultimately producing evidence-grounded long-form reports~\citep{nakano2021webgpt,lewis2020retrieval,yao2023react,schick2023toolformer}. As these agents are increasingly used in realistic research workflows, rigorous evaluation becomes essential, motivating recent benchmarks that assess long-form report generation, research-tools integration and research-process quality~\citep{du2025deepresearch,han2025deer,coelho2025deepresearchgym,wang2026deepresearcheval,ye2026miroevalbenchmarkingmultimodaldeep}. However, evaluation remains challenging: the final report is long-form and knowledge-intensive, making cited sources difficult to verify; and the execution trajectory is multi-step, open-ended, and difficult to audit, making it hard to assess whether a fluent report truly reflects sound retrieval and well-supported claims.

\begin{figure}[h!]
    \centering
    \includegraphics[width=1\linewidth]{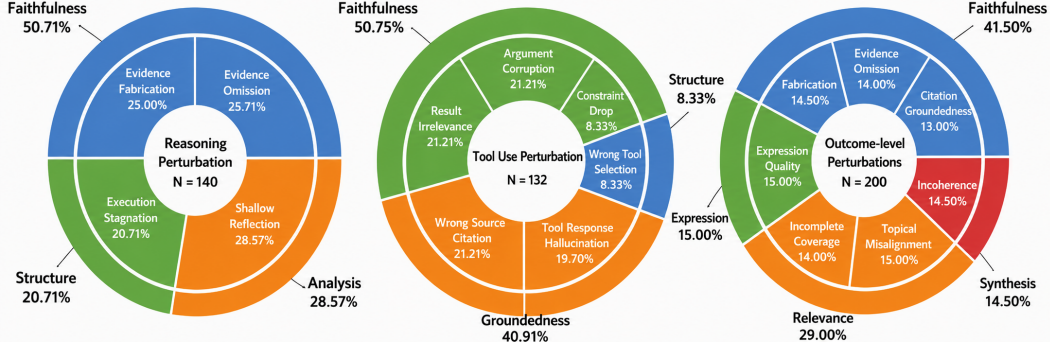}
    \vspace{-2ex}
\caption{Data distribution of \benchmark{} across reasoning-process ($N=140$), tool-use ($N=132$), and outcome-level ($N=200$) error types. The outer rings represent the high-level failure dimensions of deep research agents and their corresponding proportions, while the inner rings break each dimension down into fine-grained error types defined by our taxonomy, which is summarized from prior work (see Table~\ref{tab:process_error_taxonomy_related_work}) and further verified through case studies of natural rollouts (see Appendix~\ref{app:case_study}).}
    \label{fig:tax}
    \vspace{-3ex}

\end{figure}

Such challenges make human evaluation over full research trajectories costly and infeasible at scale, motivating LLM-as-judge as a scalable supervision paradigm for assessing report quality, tool integration, and intermediate reasoning processes~\citep{zheng2023judgingNeurIPS,liu2023geval,dubois2023alpacafarm,du2025deepresearch,coelho2025deepresearchgym,han2025deer,ye2026miroevalbenchmarkingmultimodaldeep}. Related work further uses LLM judges or reward models to supervise search behavior, step-level reasoning, and citation-aware training signals~\citep{zhan2026whyyour,shao2025seekbench,chen2026srrjudge,hu2025stepdeepresearch,zhang2026carr,shao2025drtulu}. Yet the reliability of these judges when evaluating deep research agents remains poorly understood, posing a critical meta-evaluation problem~\citep{liu2025reife}: before deploying LLM judges to supervise research agents, we must first evaluate the judges themselves.

However, existing meta-evaluation protocols are ill-suited for assessing judge reliability in deep research agent settings. Prior work validates automated judges by measuring agreement with human ratings, rankings, or pairwise preferences over model outputs~\citep{hwang2026shallow,bosse2025deepresearchbench,coelho2025deepresearchgym,li2025researchqa}. This paradigm leaves three critical gaps for evidence-based research agents:
\textbf{(1) Coarse and subjective labels.} Overall preferences indicate which output humans favor, but shed little light on which specific failures a judge detects or misses.
\textbf{(2) Absence of ground truth in open-ended tasks}. Prior meta-evaluation targets settings with verifiable answers, such as mathematics, coding, or factual QA. Deep research agents instead operate in open-ended settings with no single correct answer or canonical trajectory, making reliable labels difficult to construct for retrieval, tool use, reasoning, and synthesis.
\textbf{(3) Insufficient coverage of process-level execution.} Existing protocols assess judges against coarse human judgments over final outputs, offering limited insight into whether LLM judges can detect process-level failures such as poor evidence gathering or tool misuse.

To address these gaps, we introduce \textbf{\benchmark{}}~(\textbf{RE}liable \textbf{F}ine-grained \textbf{L}LM judge \textbf{E}valuation via \textbf{C}ontrolled in\textbf{T}ervention), a meta-evaluation benchmark targeting fine-grained failure detection of LLM judges for non-verifiable agentic execution. \benchmark{} offers three key advantages: \textbf{(1) Verifiable ground-truth labels}: instead of relying on subjective human preferences, we make controlled, localized interventions on quality-screened agent trajectories and reports, making labels objective and directly verifiable by construction. \textbf{ (2) Comprehensive and realistic failure coverage}: perturbations are drawn from a taxonomy of realistic failures spanning both process- and outcome-level errors in reasoning, tool use, evidence gathering, and synthesis. \textbf{ (3) Fine-grained diagnostic signal}: by reframing meta-evaluation as failure detection with known failure types and locations, \benchmark{} enables precise identification of judge blind spots and systematic comparison between fine-grained and holistic evaluation paradigms.

Using \benchmark{}, we evaluate various LLM-judges, including both those that perform holistic and fine-grained, step-level evaluations. Our experiments reveal major reliability gaps in current LLM judges: judges fail in different ways, and no single aggregate score captures overall reliability. Fine-grained evaluation is more effective than holistic scoring, particularly for macro-level structural failures that require cross-stage tracing. Overall, \benchmark{} exposes overlooked failure types and vulnerable components, offering guidance for improving judge prompts and protocols for more reliable agentic research systems. Our contributions are threefold:

\begin{enumerate}[leftmargin=0pt, itemindent=*, labelsep=0.5em]
    \item We introduce \benchmark{}, the first comprehensive and fine-grained meta-evaluation benchmark for assessing LLM judges in deep research agent executions traces and reports, converting judge evaluation from subjective, coarse preference matching into targeted failure detection.
    \item We construct the benchmark using controlled, localized interventions based on comprehensive error taxonomy of deep research agent, producing instances with specific failure types and verifiable ground-truth labels.
    \item We systematically study judge reliability and cost across models, failure categories, and evaluation protocols. Our findings reveal major reliability gaps in current judges and point to fine-grained judging protocols as a potential enhancement for robust evaluation pipelines for deep research agents.
\end{enumerate}

%% file: sections/method.tex
\section{\benchmark{}}
\label{sec:framework}
  \vspace{ -1ex}
\subsection{Benchmark Task Formulation}
\label{sec:testbed}
   \vspace{ -1ex}

\benchmark{} frames judge meta-evaluation as an accuracy-based preference task over
research-agent executions. Each instance pairs a reference execution
with a controlled failure-bearing alternative. A reliable judge should assign
higher quality to the reference, thereby showing sensitivity to the targeted
failure. This formulation supports both process-level evaluation of trajectories
and outcome-level evaluation of final reports, while retaining verifiable labels
for open-ended research tasks that lack a single canonical answer. We first formalize the
benchmark task and failure space in \S\ref{sec:testbed}, then describe
the four-stage construction pipeline used to build verified
clean-perturbed pairs in \S\ref{sec:pipeline}.

\textbf{Agent executions.}
An evidence-based deep research agent $\mathcal{A}$ maps an input
query $q$ to an execution $\xi=(q,\tau,y)$, where $\tau$ is the
research trajectory and $y$ is the final long-form answer. Following
ReAct~\citep{yao2023react}, the trajectory is a sequence of reasoning,
tool-call, and tool-response triples,
$\tau=\big((r_t,c_t,s_t)\big)_{t=1}^{T}$, with history
$h_{<t}=(q,r_{<t},c_{<t},s_{<t})$. At each step, the agent generates
$r_t=\mathcal{A}_{\mathrm{reason}}(h_{<t})$ and selects a tool call
$c_t=\mathcal{A}_{\mathrm{tool}}(h_{<t},r_t)$. Each tool call
$c_t=(u_t,\theta_t)$ specifies a tool $u_t$ from the available tool set
$\mathcal{U}$ and its arguments $\theta_t$. The tool returns a response
$s_t=\mathcal{E}(c_t)$. After completing the trajectory, the final answer is
produced as $y=\mathcal{A}_{\mathrm{ans}}(q,\tau)$.

\textbf{Failure space.}
We partition the failure space as
$\mathcal{F}=\mathcal{F}_{\mathrm{proc}}\cup\mathcal{F}_{\mathrm{out}}$.
Process-level failures $\mathcal{F}_{\mathrm{proc}}$ arise within the
trajectory $\tau$, including errors in reasoning, tool calls, and the
use or interpretation of tool responses. Outcome-level failures
$\mathcal{F}_{\mathrm{out}}$ arise in the final answer $y$. We derived
and adapted the full error taxonomy from prior work on long-form QA,
deep research agents, and agent evaluation~\citep{zhu2025where,
zhan2026whyyour, shao2025seekbench, chen2026srrjudge}. Details of the
our error taxonomy distribution can be found in Figure \ref{fig:tax} and its relation to existing schemes are given in
Appendix~\ref{app:taxonomy_related_work}.

\begin{figure}[t!]
    \centering
    \includegraphics[width=1\linewidth]{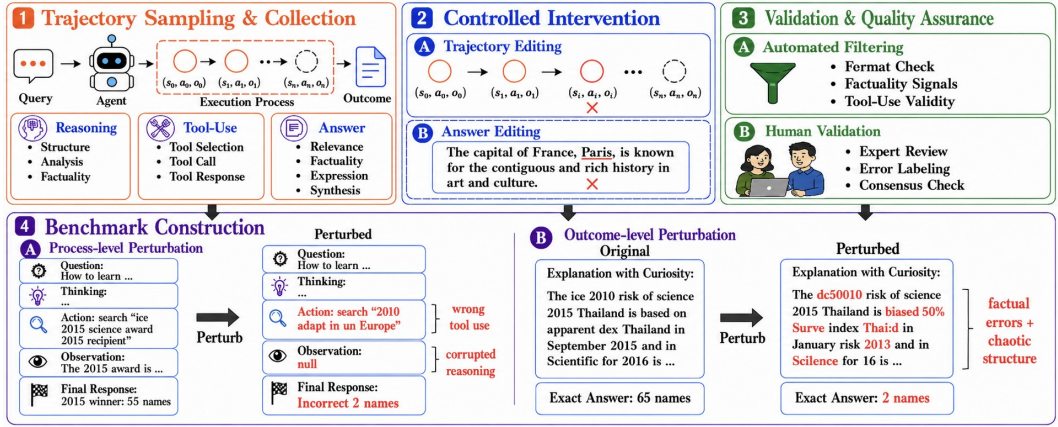}
    \caption{Overview of the benchmark construction pipeline of \benchmark{}, which collects agent trajectories, applies controlled perturbations to reasoning, tool use, and answers, and validates the resulting samples through automated filtering and human review.}
    \vspace{ -2ex}
    \label{fig:framework}
\end{figure}

\textbf{Benchmark instances.}
Given a quality-screened agent execution
$\xi^\star=(q,\tau^\star,y^\star)$ and a target failure type
$f\in\mathcal{F}$, a perturbation operator $\Pi_f$ produces
a corrupted execution $\tilde{\xi}=\Pi_f(\xi^\star)$ that contains
$f$ and differs from $\xi^\star$ only at a designated edit site. Each
benchmark instance $b_i\in\mathcal{B}$ consists of a verified
reference-corrupted execution pair, a failure label, and an edit site $\ell_i$:
\[
    \mathcal{B}
    =
    \big\{
    b_i=(\xi_i^\star,\tilde{\xi}_i,f_i,\ell_i)
    \big\}_{i=1}^{M},
    \qquad
    \tilde{\xi}_i=\Pi_{f_i}(\xi_i^\star).
\]
Here $\ell_i$ is a trajectory step $t\in\{1,\dots,T_i\}$ for
process-level perturbations and a contiguous answer chunk for
outcome-level perturbations. Since $\xi_i^\star$ is verified to be free
of $f_i$ and $\tilde{\xi}_i$ to contain it, each instance provides
ground truth for judge meta-evaluation. The edit-site metadata also
supports localization analysis, scored by step-level exact match for
trajectories and chunk-level overlap for answers; see
Section~\ref{sec:experiments}.

\textbf{Judge interfaces.}
A judge $\mathcal{J}$ is evaluated by whether it identifies or prefers
the reference execution over its failure-bearing counterpart. We support three
interfaces commonly used in evaluation and training.

\textit{Scalar judging.} A scalar (or pointwise) judge assigns a quality score
$S_{\mathcal{J}}(\xi)\in\mathbb{R}$, as in reward modeling or
score-based filtering. For a reference-corrupted pair, we define the
score gap and success indicator as
\[
    \Delta_{\mathcal{J}}(\xi^\star,\tilde{\xi})
    =S_{\mathcal{J}}(\xi^\star)-S_{\mathcal{J}}(\tilde{\xi}),
    \qquad
    z_{\mathcal{J}}(\xi^\star,\tilde{\xi})
    =\mathbb{I}\!\left[\Delta_{\mathcal{J}}(\xi^\star,\tilde{\xi})>\epsilon\right].
\]
We use $\epsilon=0$ as the default margin throughout the paper.

\textit{Pairwise judging.} A pairwise judge directly compares two
executions, matching preference-learning settings such as DPO-style
training, \cite{dpo} and returns
$P_{\mathcal{J}}(\xi^\star,\tilde{\xi})\in\{\xi^\star,\tilde{\xi},\mathrm{tie}\}$.
It succeeds when
\[
    z_{\mathcal{J}}(\xi^\star,\tilde{\xi})
    =
    \mathbb{I}\!\left[
    P_{\mathcal{J}}(\xi^\star,\tilde{\xi})=\xi^\star
    \right].
\]

\textit{Ranking judging.} A ranking judge selects the best execution
from a candidate set, corresponding to Best-of-\(N\) inference-time scaling or
reranking. Let $\mathcal{P}$ denote a set of perturbation types, each
producing a candidate $\tilde{\xi}_a$ for $a\in\mathcal{P}$. The judge
sees
$\mathcal{C}=\{\xi^\star\}\cup\{\tilde{\xi}_a:a\in\mathcal{P}\}$,
selects $T_{\mathcal{J}}(\mathcal{C})\in\mathcal{C}$, and succeeds when
\[
    z_{\mathcal{J}}(\mathcal{C})
    =
    \mathbb{I}\!\left[
    T_{\mathcal{J}}(\mathcal{C})=\xi^\star
    \right].
\]

\subsection{Benchmark Construction Pipeline}
\label{sec:pipeline}

We instantiate the perturbation operators $\Pi_f$ through a four-stage
pipeline: taxonomy construction, reference screening, controlled intervention, and
combined automated filtering and human validation.

\textbf{Taxonomy construction.}
We construct a failure space $\mathcal{F}$ by synthesizing  categories from prior
work on long-form QA, deep research benchmark, and agentic trajectory supervision \citep{zhu2025where,zhong2025complexfuncbench,chen2026srrjudge,shao2025seekbench}.
Existing taxonomies
typically emphasize either final-answer quality or trajectory behavior
in isolation; ours unifies both views and is the basis for the
process/outcome partition above. To verify that the taxonomy reflects
real agent behavior rather than an a priori list, we sample natural
rollouts on held-out queries and, under model-assisted and human
review, map each observed failure either to a category in $\mathcal{F}$
or to an out-of-taxonomy bucket. Case studies can be found
in Appendix~\ref{app:case_study}.

\textbf{Reference screening.}
We draw candidate reference executions from strong agent rollouts. For each target failure type \(f\),
we only require that the selected reference does not already contain \(f\) at
the chosen edit (i.e. the step or chunk to edit). Candidate references are screened using automatic checks
for schema validity, English language content, and usable trajectory or answer
structure, followed by targeted validation for the absence of \(f\).

\textbf{Controlled Intervention.}
For each failure type $f\in\mathcal{F}$, we define a perturbation
operator $\Pi_f$ implemented as an LLM-based editor. Starting from
clean seeds $\xi^\star$ obtained from strong agent rollouts, we use LLM along with human supervision to
pre-filter to ensure $f$ is absent in the seed, then apply
\[
    \tilde{\xi}=\Pi_f(\xi^\star)
    =\mathrm{Edit}_{\theta}\!\big(\xi^\star,\,f,\,d_f,\,\ell;\,p_f\big),
\]
where $\mathrm{Edit}_{\theta}$ denotes an LLM editor with parameters
$\theta$, $d_f$ is a natural-language definition of the failure type,
$\ell$ is the target edit site sampled from candidate sites in
$\xi^\star$, and $p_f$ is a type-specific perturbation prompt. Each
operator targets either the trajectory $\tau^\star$ (for
$f\in\mathcal{F}_{\mathrm{proc}}$) or the answer $y^\star$ (for
$f\in\mathcal{F}_{\mathrm{out}}$); concrete worked examples are listed in Appendix~\ref{app:case_study}.

Following adversarial comparison benchmarks such as
LLMBar~\citep{zeng2024llmbar}, we constrain edits to be
\textit{localized}, \textit{plausible}, and \textit{minimal}: a
perturbation should introduce $f$ at $\ell$ while preserving fluency,
coherence, and all content outside $\ell$. For trajectories, this means
that surrounding steps and their observations are left unchanged and
the edited step remains syntactically well-formed; for answers, it
means that only the targeted chunk is rewritten. This discipline
ensures that judge success depends on detecting $f$ rather than
exploiting superficial artifacts such as length, formatting, or
stylistic drift.

\textbf{Automated filtering and human validation.}
Every original-perturbed pair is screened by automated filters that
remove pairs with no substantive change, malformed outputs, non-English
text, formatting artifacts, or invalid input-output structure for the
target judge interface. Pairs that pass the filter proceed to a human validation step.
Two annotators with graduate-level expertise in NLP independently
verify three conditions for every pair: $\tilde{\xi}$ contains the
target failure $f$, $\xi^\star$ does not, and the perturbation
introduces no major unintended failures. Annotators completed a
calibration round on a held-out development sample before the main
study, and disagreements on the main study were resolved through
adjudicated discussion. We obtain an inter-annotator agreement of
$\kappa=0.86$, indicating substantial agreement. 
Final dataset statistics are illustrated in Figure \ref{fig:tax} and report in 
in Table~\ref{tab:dataset_statistics} in Appendix~\ref{app:dataset_statistics}.

%% file: sections/experiment.tex
\section{Experiments}\label{sec:experiments}

We design our experiments to answer the following key research questions:

\textbf{RQ1: Model capability.} How do different judge models perform in detecting various fine-grained failure modes, and how do open-weight models compare with frontier closed-source models?
\textbf{RQ2: Evaluation protocol.} How do judging protocols, including holistic versus fine-grained evaluation, rubric guidance, and explicit reasoning, affect the judge's reliability?
\textbf{RQ3: Judge blind spots.} Which process-level and outcome-level failure types are systematically missed by LLM judges, and how do these blind spots depend on evaluation granularity?
\textbf{RQ4: best-of-\(N\) and cost-performance trade-off.} Can LLM judges identify the verified original execution among multiple failure-bearing alternatives (a useful setup for best-of-N inference-time scaling), and which protocol choices provide the best reliability-cost trade-off?

\subsection{Experimental Setup}

\textbf{Evaluation Protocols.}
We formulate judge reliability as an accuracy-based preference task: given a verified reference execution and failure-bearing alternatives, the judge should prefer the reference. We evaluate two targets: the agent's execution process and its final output. Process-level evaluation assesses trajectories, distinguishing \emph{reasoning behavior} (e.g., planning, reflection, and evidence use) from \emph{tool-use behavior} (e.g., tool selection, argument construction, and response interpretation), while outcome-level evaluation assesses the final report.

We vary three protocol axes: (i) \emph{judging granularity}, comparing holistic judgments over full trajectories or reports with fine-grained judgments over localized steps or chunks; (ii) \emph{comparison format}, comparing pointwise independent scoring with pairwise direct comparison; and (iii) \emph{prompting format}, comparing rubric-based judgments with non-rubric overall judgments. For pairwise evaluation, we use a swapped-order design to mitigate position bias \citep{liu2025reife}. Additional prompt details are provided in Appendix~\ref{app:implementation_details}.

\textbf{Judge Models.}
We evaluate a wide array of LLM judges covering both open-weight and proprietary models 
.
Such judges are increasingly used beyond offline evaluations, such as  best-of-\(N\) selection and RL-style training signals \cite{zhou2025evaluating,malik2026rewardbench}. 
The open-weight judges include Qwen3-8B, Qwen3-32B, and Qwen3-235B-A22B~\citep{yang2025qwen3}, Llama-3.1-70B~\citep{grattafiori2024llama}, Gemma3-27B~\citep{gemmateam2025gemma3}, and GPT-OSS-120B~\citep{agarwal2025gptoss}. 
The proprietary judges include Gemini-2.0-Flash~\citep{google2025gemini20flash}, Gemini-2.5-Flash~\citep{google2025gemini25flash}, Gemini-3.1-Pro~\citep{google2026gemini31pro}, GPT-5.3-Codex~\citep{openai2026gpt53codex}, GPT-5.4~\citep{openai2026gpt54}, GPT-5-mini~\citep{openai2025gpt5mini}, Claude-Haiku-4.5~\citep{anthropic2025haiku45}, and Claude-Opus-4.7~\citep{anthropic2026opus47}.

\textbf{Benchmark Instances.}
Our benchmark draws on different sources for process-level and outcome-level perturbations. 
For process-level evaluation, we use clean agent trajectories from two trace sources: cleaned DR.TULU~\citep{shao2025drtulu} and Tongyi DeepResearch~\citep{team2025tongyi}.\footnote{
We do not assume that source trajectories are globally error-free. 
They are used as reference executions after screening and validation for the target failure type: the reference must not contain the target failure at the selected edit site, while the edited alternative must contain that failure and preserve the surrounding trajectory. 
This paired counterfactual design controls for residual imperfections shared by both executions and tests whether judges are sensitive to the controlled localized degradation.
} 
These traces provide the reasoning and tool-use steps used to construct process-level perturbations. 
For outcome-level evaluation, we use final reports from cleaned DR.TULU~\citep{shao2025drtulu} and English final answers sampled from DeepResearch Bench~\citep{du2025deepresearch}. 
All instances are normalized into a shared format containing the user question, final answer, and trajectory steps when available.

\textbf{Metrics.}
We use \textbf{accuracy} as the primary metric, consistent with reward-model
and judge meta-evaluation benchmarks \citep{malik2026rewardbench, zhou2025evaluating}, and following the scalar-judging
success criterion defined in Section~\ref{sec:testbed}. A judge is correct on
a pair $(\xi_i^\star,\tilde{\xi}_i)$ if the original execution receives a
strictly higher final score than the perturbed execution. I.e., $
\mathrm{Acc}
=
\frac{1}{N}
\sum_{i=1}^{N}
\mathbb{I}\!\left[
    S_{\mathcal{J}}(\xi_i^\star)
    >
    S_{\mathcal{J}}(\tilde{\xi}_i)
\right].$

For non-rubric scoring, $S_{\mathcal{J}}$ is the judge's direct overall score.
For rubric scoring, $S_{\mathcal{J}}(x)=\frac{1}{K}\sum_{k=1}^{K}s_{\mathcal{J},k}(x)$,
where $s_{\mathcal{J},k}(x)\in\{1,\ldots,n\}$ is the score for rubric dimension
$k$. We report accuracy overall and by failure type.

\definecolor{overallblue}{RGB}{232,242,255}

\begin{table}[t]
\centering
\caption{
Detection accuracy for process- and outcome-level evaluations with pointwise judges. Values are
percentages with \% omitted. Abbreviations: AN = Analysis, ST = Structure, OV = Overall, FI = Faithfulness, GR = Groundedness,
RE = Relevance, EX = Expression, SY = Synthesis. Bold and underline mark the
best and the runner-up.
}
\label{tab:main_table}
\resizebox{\textwidth}{!}{%
\begin{tabular}{lrrr>{\columncolor{overallblue}}crrr>{\columncolor{overallblue}}crrrr>{\columncolor{overallblue}}c}
\toprule
\multirow{3}{*}{\textbf{Model}}
& \multicolumn{4}{c}{\textbf{Process-level: Reasoning}}
& \multicolumn{4}{c}{\textbf{Process-level: Tool Use}}
& \multicolumn{5}{c}{\textbf{Outcome-level: Report Quality}} \\
\cmidrule(lr){2-5} \cmidrule(lr){6-9}  \cmidrule(lr){10-14}
& AN & ST & FI & OV
& ST & FI & GR & OV
& RE & FI & EX & SY & OV \\
\midrule

\multicolumn{14}{l}{\textbf{Open-source Models}} \\
\midrule

Qwen3-8B
& 0.0 & 3.4 & 0.0 & 0.7
& 0.0 & 7.5 & 0.0 & 3.8
& 5.2 & 9.6 & 26.7 & 34.5 & 14.5 \\

Qwen3-32B
& 0.0 & 0.0 & 0.0 & 0.0
& 0.0 & 0.0 & 3.7 & 1.5
& \underline{34.5} & \underline{34.9} & \textbf{50.0} & \underline{51.7} & \underline{39.5} \\

Gemma3-27B
& 5.0 & 10.3 & 7.0 & 7.1
& 0.0 & 1.5 & 1.9 & 1.5
& 20.7 & 19.3 & 6.7 & 34.5 & 20.0 \\

Llama3.1-70B
& 0.0 & 0.0 & 0.0 & 0.0
& 9.1 & 1.5 & 1.9 & 2.3
& 6.9 & 7.2 & 20.0 & 10.3 & 9.5 \\

Qwen3-235B-a22B
& \underline{30.0} & \underline{24.1} & \underline{22.5} & \underline{25.0}
& \underline{18.2} & \underline{19.4} & \underline{22.2} & \underline{20.5}
& 17.2 & 32.5 & 30.0 & 27.6 & 27.0 \\

GPT-OSS-120B
& \textbf{57.5} & \textbf{48.3} & \textbf{38.0} & \textbf{45.7}
& \textbf{36.4} & \textbf{28.4} & \textbf{24.1} & \textbf{27.3}
& \textbf{43.1} & \textbf{43.4} & \underline{43.3} & \textbf{65.5} & \textbf{46.5} \\

\midrule
\multicolumn{14}{l}{\textbf{Closed-source Models}} \\
\midrule

Gemini-2.0-Flash
& 2.6 & 3.4 & 6.0 & 4.5
& 0.0 & 4.5 & 1.9 & 3.0
& 3.4 & 6.0 & 3.3 & 3.4 & 4.5 \\

Gemini-2.5-Flash
& \underline{33.3} & 31.0 & 32.8 & \underline{32.6}
& 27.3 & 19.4 & 20.4 & 20.5
& 22.4 & 18.1 & 30.0 & 41.4 & 24.5 \\

Gemini-3.1-Pro
& 20.0 & 31.0 & 23.5 & 24.1
& \underline{54.5} & 28.4 & 22.2 & 28.0
& \textbf{41.4} & 30.1 & 23.3 & 31.0 & 32.5 \\

Claude-Haiku-4.5
& 15.0 & 20.7 & 12.7 & 15.0
& 63.6 & \underline{41.8} & 40.7 & 43.2
& 32.8 & 31.3 & 30.0 & 37.9 & 32.5 \\

Claude-Opus-4.7
& 15.0 & 20.7 & 21.1 & 19.3
& \underline{81.8} & 49.3 & \textbf{51.9} & \underline{53.0}
& 29.3 & 37.3 & 26.7 & 48.3 & 35.0 \\

GPT-5.4
& \textbf{40.0} & \underline{41.4} & \underline{33.8} & \textbf{37.1}
& \textbf{90.9} & \underline{55.2} & 37.0 & 50.8
& \underline{39.7} & 33.7 & \underline{36.7} & 48.3 & 38.0 \\

GPT-5-mini
& 30.0 & \textbf{48.3} & \textbf{36.6} & \textbf{37.1}
& 36.4 & 28.4 & \underline{42.6} & 34.8
& 34.5 & \underline{44.6} & \textbf{43.3} & \textbf{58.6} & \underline{43.5} \\

GPT-5.3-codex
& 30.0 & 37.9 & 19.7 & 26.4
& 63.6 & \textbf{59.7} & \underline{46.3} & \textbf{54.5}
& \underline{39.7} & \textbf{51.8} & \textbf{43.3} & \underline{55.2} & \textbf{47.5} \\
\bottomrule
\end{tabular}
}
\vspace{-3ex}
\end{table}

\subsection{Model Capability (RQ1)} 

Table~\ref{tab:main_table} shows that judge performance remains low across process-level
reasoning, tool-use accuracy, and outcome-level evaluations. The results reveal the following main findings.

\noindent\textbf{\textit{Existing judges remain unreliable.}}
Even the best overall scores are far from reliable: 45.7\% for
reasoning, 54.5\% for tool use, and 47.5\% for report quality. Performance
also varies widely across model families and evaluation targets. Smaller
open-weight judges perform poorly in most settings, while larger open-weight
and proprietary models are more competitive but still unreliable.

\noindent\textbf{\textit{Judge reliability is failure-type dependent.}}
Tool-use ``structure'' errors are generally easier for several frontier models to detect,
whereas ``groundedness'' and ``faithfulness'' failures remain substantially more
challenging. At the outcome level, models also differ in whether they are more sensitive
to relevance, faithfulness, expression, or synthesis failures. This heterogeneity indicates
that aggregate accuracy alone can obscure important differences in what judges can and
cannot detect.

\noindent\textbf{\textit{Agent-oriented judges are strongest overall.}}
The strongest overall results come from \texttt{GPT-5.3-codex}, which achieves
the best process-level tool-use accuracy of 54.5\% and the best outcome-level
report accuracy of 47.5\%. This suggests that models optimized for agentic coding
and tool-oriented tasks may be well suited to evidence-based judge evaluation,
though this advantage comes with higher inference cost and requires further
controlled study.

\subsection{Evaluation Protocol Comparison (RQ2)}

We next study how protocol choices affect judge reliability. Because exhaustive
fine-grained protocol sweeps are expensive, we run these comparisons on a
representative subset of judges spanning open-weight and proprietary models. For
evaluation granularity, we compare holistic judging over the full trajectory or
report with fine-grained judging over localized trajectory steps or answer
chunks. Table~\ref{tab:scale_effect_by_setting} reports
$\Delta_{\text{scale}}$, the accuracy difference between fine-grained and
holistic judging, across process- and outcome-level settings.

\noindent\textbf{\textit{Fine-grained evaluation improves over holistic judging.}}
Fine-grained judging consistently improves detection accuracy across models,
evaluation levels, and rubric settings. The gains are substantial in both
\textit{process-level} and\textit{ outcome-level} evaluation, with $\Delta_{\text{scale}}$ reaching
over 30 points in several settings. This suggests that localized evaluation helps
judges identify errors that may be diluted under holistic scoring, whether they
appear in intermediate reasoning trajectories or in final reports. Overall, the
results show that granularity is a robust protocol effect, and we next ask
whether explicit rubric dimensions provide an additional source of judge
reliability.

\definecolor{deltagreen}{RGB}{92,128,112}
\newcommand{\gain}[1]{\textcolor{deltagreen}{\footnotesize(#1)}}

\begin{table*}[t]
\centering
\caption{
Effect of evaluation granularity across process- and outcome-level settings.
$\Delta_{\text{scale}}$ denotes the difference between fine-grained and holistic
detection accuracy, measured in percentage points.
}
\label{tab:scale_effect_by_setting}
\footnotesize
\setlength{\tabcolsep}{4.3pt}
\begin{tabular}{lrrrrrrrrrrrr}
\toprule
\multirow{3}{*}{Model}
& \multicolumn{6}{c}{Process-Level}
& \multicolumn{6}{c}{Outcome-Level} \\
\cmidrule(lr){2-7} \cmidrule(lr){8-13}
& \multicolumn{3}{c}{Rubric}
& \multicolumn{3}{c}{No-Rubric}
& \multicolumn{3}{c}{Rubric}
& \multicolumn{3}{c}{No-Rubric} \\
\cmidrule(lr){2-4} \cmidrule(lr){5-7}
\cmidrule(lr){8-10} \cmidrule(lr){11-13}
& Hol. & FG & $\Delta_{\text{scale}}$
& Hol. & FG & $\Delta_{\text{scale}}$
& Hol. & FG & $\Delta_{\text{scale}}$
& Hol. & FG & $\Delta_{\text{scale}}$ \\
\midrule

Qwen3-8B
& 0.7 & 20.0 & \gain{+19.3}
& 3.6 & 25.7 & \gain{+22.1}
& 14.5 & 34.0 & \gain{+19.5}
& 10.5 & 25.4 & \gain{+14.9} \\

Qwen3-32B
& 0.0 & 21.4 & \gain{+21.4}
& 0.0 & 34.3 & \gain{+34.3}
& 39.5 & 45.5 & \gain{+6.0}
& 14.5 & 37.6 & \gain{+23.1} \\

GPT-5.4
& 37.1 & 55.7 & \gain{+18.6}
& 22.9 & 56.4 & \gain{+33.5}
& 38.0 & 55.3 & \gain{+17.3}
& 32.0 & 34.8 & \gain{+2.8} \\

Gemini-3.1 Pro
& 24.1 & 55.4 & \gain{+31.3}
& 25.0 & 55.7 & \gain{+30.7}
& 32.5 & 56.9 & \gain{+24.4}
& 12.5 & 23.2 & \gain{+10.7} \\

\bottomrule
\end{tabular}%
\vspace{0.5ex}
\begin{flushleft}
\vspace{-1mm}
\footnotesize
Values are detection accuracy percentages. Hol. = holistic; FG = fine-grained. 
\vspace{-4ex}
\end{flushleft}
\end{table*}

\begin{figure*}[t]
    \centering

    \begin{subfigure}[t]{0.52\textwidth}
        \centering
        \includegraphics[width=\linewidth]{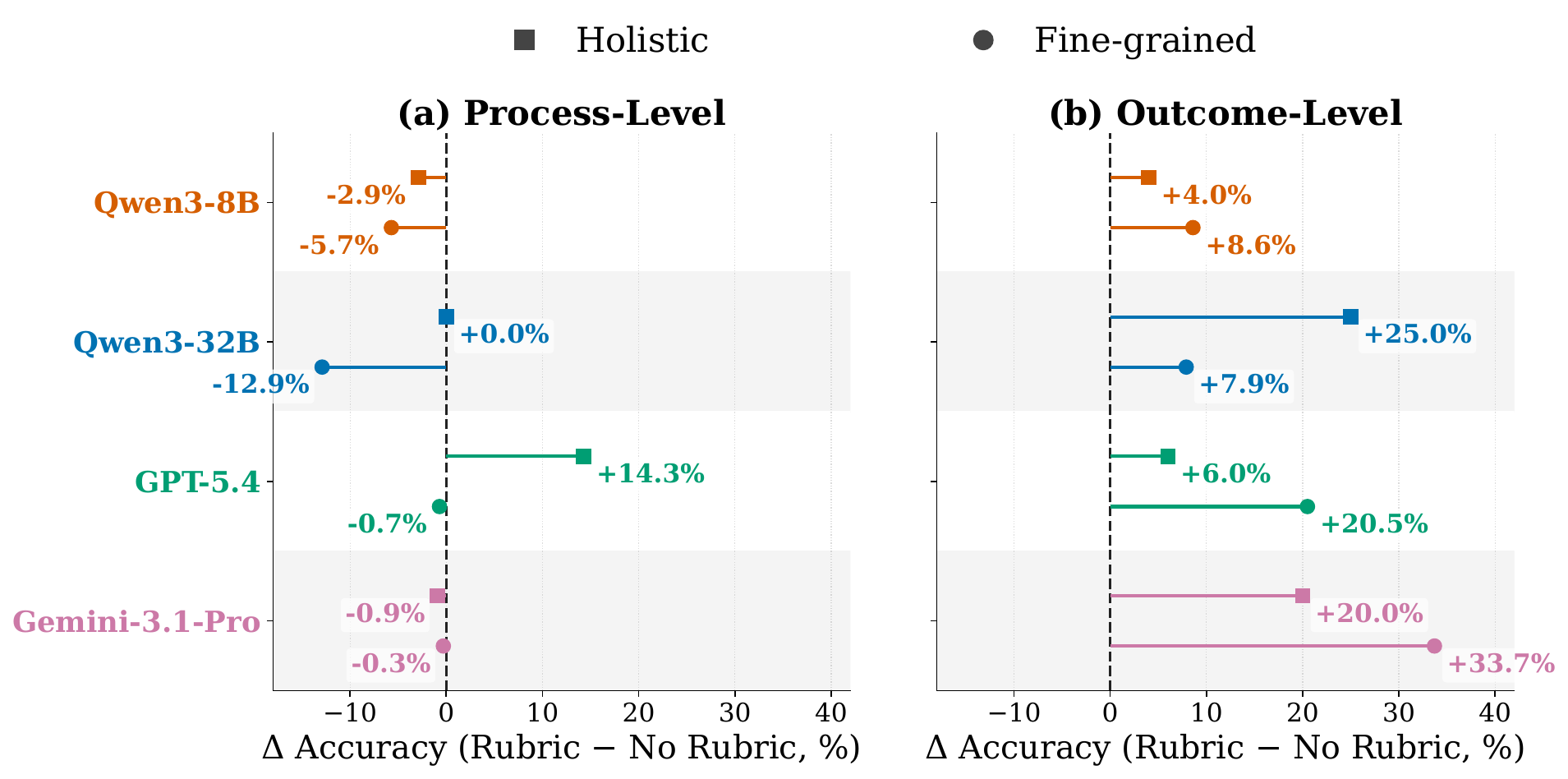}
        \caption{Rubric benefit across models and granularities.}
        \label{fig:rubric_delta}
    \end{subfigure}
    \hfill
    \begin{subfigure}[t]{0.47\textwidth}
        \centering
       \includegraphics[width=\linewidth]{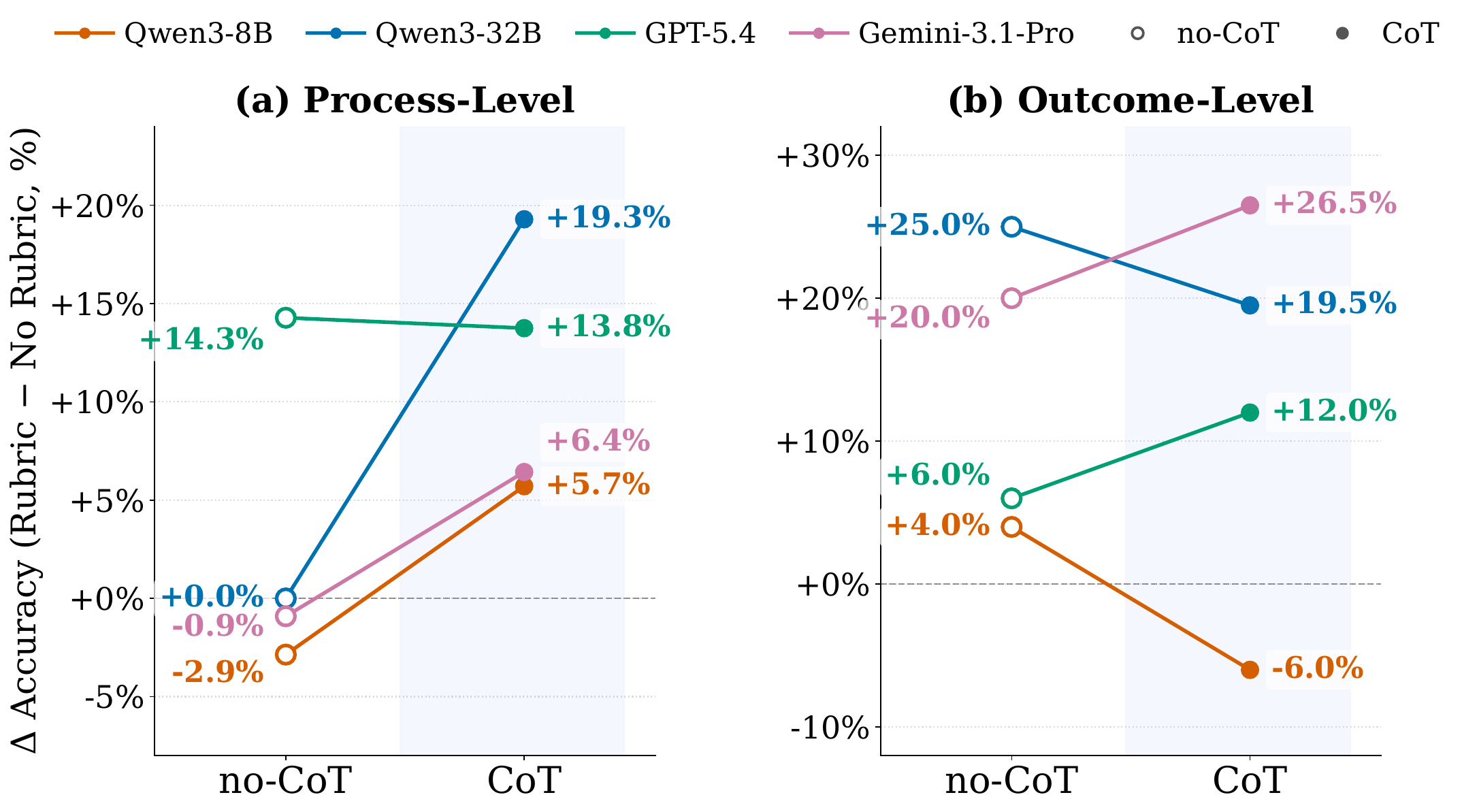}
        \caption{CoT reasoning effect with rubrics.}
        \label{fig:cot_rubric_interaction}
    \end{subfigure}
\vspace{-0.5ex}
    \caption{
    Effects of rubric-guided evaluation and chain-of-thought reasoning on perturbation detection accuracy.
    $\Delta$ denotes Rubric $-$ No-Rubric accuracy in percentage points.
    }
     \vspace{-2ex}
    \label{fig:rubric_cot_combined}
\end{figure*}

\noindent\textbf{\textit{Rubric effects are context-dependent, motivating dynamic instance-based rubrics.}}
Figure~\ref{fig:rubric_delta} shows that dimension-wise rubric scoring is not a
uniform improvement. At the \textit{outcome level}, rubrics consistently improve detection
accuracy across all selected models, with the largest gain reaching $+33.7$
points for \texttt{Gemini-3.1-Pro} under fine-grained judging. This suggests that
final-report failures align relatively well with explicit scoring dimensions,
allowing rubrics to expose localized factuality, evidence-use, or citation errors
that overall scores may overlook. In contrast, \textit{process-level} effects are mixed
and sometimes negative, especially for \textit{weaker judges}. For process evaluation,
rubrics turn a single overall decision into a more demanding task: reading long
trajectories, locating cross-step evidence, separating nearby error dimensions,
and calibrating multiple scores. When the judge lacks sufficient long-context
reasoning or scoring stability, this extra structure can become noise rather than
guidance. Overall, rubric scoring is most useful when the judge is strong enough
to apply it reliably, motivating more adaptive, instance-specific rubrics for
process-level evaluation.

 \vspace{-0.5ex}

\textbf{\textit{CoT helps only when judges can effectively leverage rubrics.}}
We next examine whether chain-of-thought (CoT) prompting further improves rubric-guided judging.
Figure~\ref{fig:cot_rubric_interaction} compares rubric gains with and without CoT: point height represents the rubric benefit, and slope indicates how this benefit changes after adding CoT.

The results show that CoT amplifies rubric gains only selectively.
It is most helpful for \textit{stronger judges} in \textit{outcome-level} evaluation, where final-report errors align well with rubric dimensions such as relevance, factuality, expression, and synthesis.
For \textit{process-level} evaluation, however, the effect is more mixed, as judges must track reasoning, tool use, and evidence flow across multiple steps.
Thus, CoT is better characterized as a capability-dependent complement to rubric-based evaluation, rather than a uniformly effective intervention.

\subsection{Blind Spots across Error Taxonomy (RQ3)}

\begin{figure*}[t]
    \centering
    \includegraphics[width=1\linewidth]{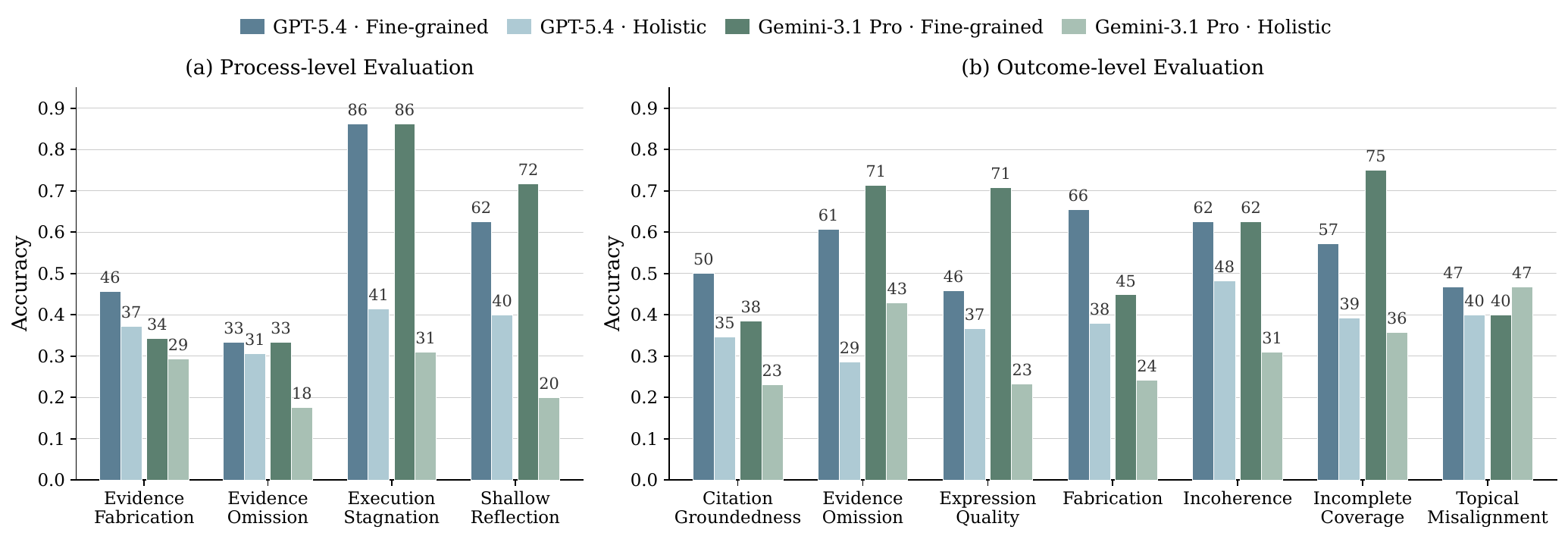}
    \caption{
    Failure detection accuracy across process-level and outcome-level perturbation types. 
    Results are shown for GPT-5.4 and Gemini-3.1 Pro under fine-grained and holistic judging.
    }
    \vspace{-2ex}
    \label{fig:stepwise_vs_holistic_accuracy}
\end{figure*}

\textbf{\textit{Fine-grained judging surfaces local errors, whereas holistic judging captures context-dependent global failures.}}
Figure~\ref{fig:stepwise_vs_holistic_accuracy} compares GPT-5.4 and Gemini-3.1 Pro across perturbation types under fine-grained and holistic judging. The results reveal granularity-dependent blind spots: failures that are salient at the step or span level may be obscured in a full trajectory, while failures that depend on broader context may only emerge when the entire response is evaluated.

\textbf{Fine-grained judging} is most effective for localized failures because it makes the perturbed step or answer span directly visible. This helps identify local process failures such as \textit{execution stagnation}, as well as local outcome failures such as \textit{evidence omission}, \textit{expression quality}, and \textit{incomplete coverage}. Under holistic judging, these signals can be diluted as they are embedded within a longer reasoning trajectory or report.\textbf{ Holistic judging}, in contrast, is better suited to failures that require global context, accumulated evidence, or overall task intent to detect. These include \textit{shallow reflection} and \textit{topical misalignment},
which may not be obvious from any single step but become clearer when the
response is evaluated as a whole.

\subsection{Best-of-N Selection and Cost Trade-offs (RQ4)}
\label{sec:best_of_n}

\paragraph{Best-of-\(N\) Metric.}
Beyond single-pair discrimination, many evaluation and deployment pipelines use
judges for best-of-\(N\) inference-time selection: the system generates multiple
candidate executions and selects the candidate with the highest judge score. We
model this setting by grouping each verified execution $\xi_i^\star$ with its
failure-bearing alternatives,
$\mathcal{C}_i=\{\xi_i^\star\}\cup\{\tilde{\xi}_{i,f}:f\in\mathcal{F}_i\}$,
where $\mathcal{F}_i$ denotes the failure types instantiated for that execution.
A group is correct only if the verified reference receives the highest judge
score:
$
    S_{\mathcal{J}}(\xi_i^\star)
    >
    \max_{f\in\mathcal{F}_i}
    S_{\mathcal{J}}(\tilde{\xi}_{i,f}).
$
Best-of-\(N\) accuracy is the fraction of groups satisfying this condition.

\begin{figure*}[t]
    \centering

    \begin{subfigure}[t]{0.59\textwidth}
        \centering
        \includegraphics[width=\linewidth]{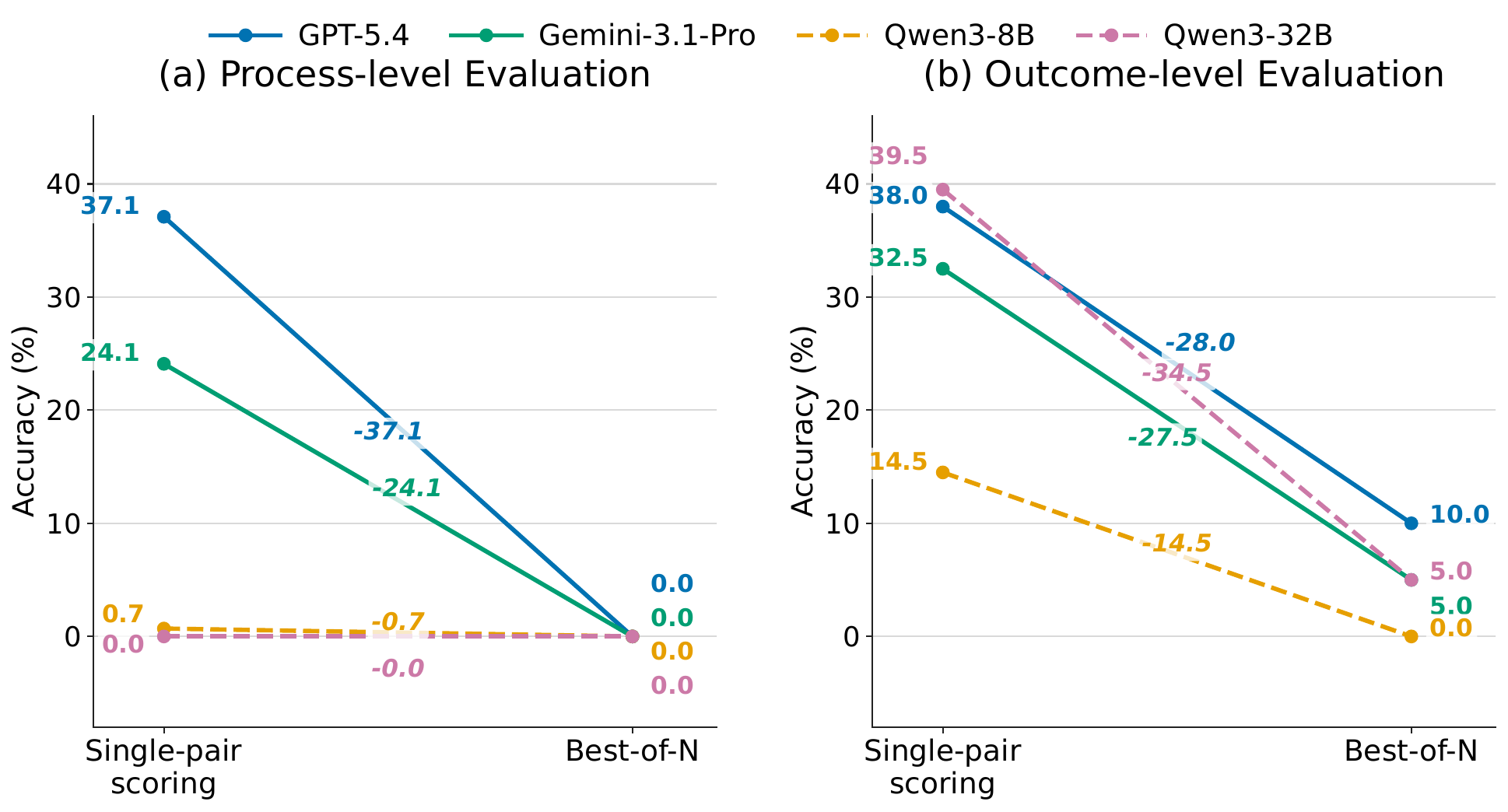}
        \caption{Best-of-\(N\) selection accuracy}
        \label{fig:reranking_detection_accuracy}
    \end{subfigure}
    \hfill
    \begin{subfigure}[t]{0.38\textwidth}
        \centering
        \includegraphics[width=\linewidth]{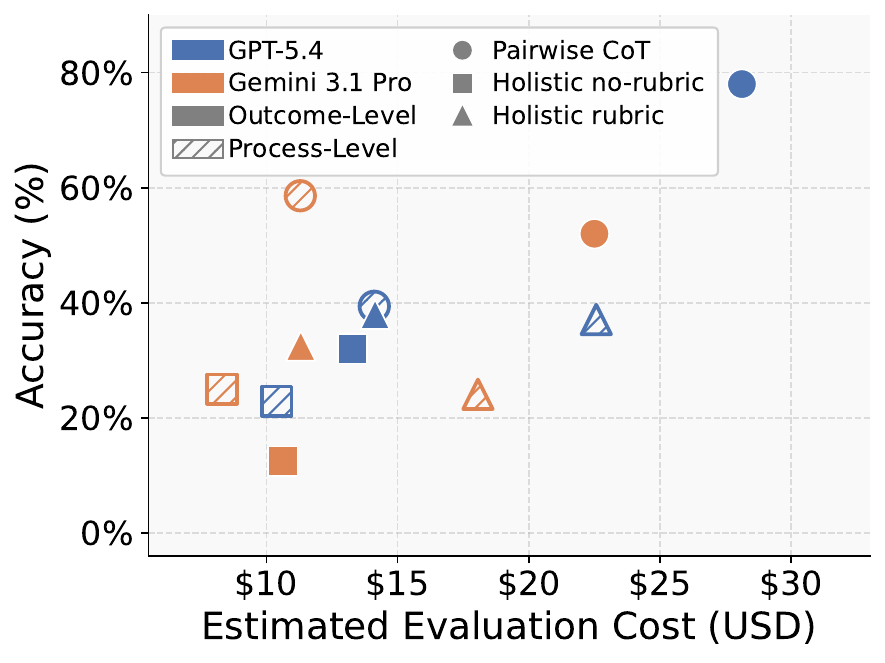}
        \caption{Cost--performance trade-off}
        \label{fig:cost_vs_accuracy}
    \end{subfigure}

    \caption{
    Judge reliability across evaluation settings.
    (a) Best-of-\(N\) selection accuracy. Single-pair scoring evaluates each reference--alternative pair independently, while Best-of-\(N\) selection requires the judge to select the verified reference among 4--7 failure-bearing alternatives.
    (b) Estimated total evaluation cost versus detection accuracy for closed-source judge settings, computed over the full benchmark by multiplying the measured input/output token counts by each model's API pricing.
    }
    \vspace{-1ex}
    \label{fig:reranking_cost}
\end{figure*}

\textbf{Best-of-\(N\) Selection Gap.}
Figure~\ref{fig:reranking_detection_accuracy} shows a clear
\emph{best-of-\(N\) selection gap}: accuracy drops when judges must select the
verified reference from a candidate set instead of scoring a single
reference-alternative pair. The drop is largest at the \emph{process level},
where selection requires comparing multiple long trajectories and tracking
distributed reasoning, tool-use, and evidence-flow failures across candidates.
The \emph{outcome-level} setting is less affected, likely because final reports
provide a more compact and directly comparable evaluation target. These results
indicate that judge scores are less reliable for best-of-\(N\) selection than
for isolated pairwise discrimination, especially when candidate quality differs
in trajectory-level behavior.

\textbf{Cost-Performance Trade-off.}
Finally, we analyze the trade-off between judge reliability and evaluation cost across closed-source judge settings.
Figure~\ref{fig:cost_vs_accuracy} shows a general \emph{positive cost-performance trend}: higher-cost settings usually obtain higher detection accuracy.
The strongest accuracies come from more expensive pairwise CoT configurations, although the gains are not determined by cost alone.
\emph{Process-level} evaluation is less cost-effective, as long trajectories increase token cost while remaining harder to judge.
Taken together, the results indicate that reliable judge evaluation requires balancing model strength and protocol design rather than simply choosing the most expensive setting.

%% file: sections/related_works.tex
\section{Related Works}
\vspace{-1ex}
\textbf{Evidence-Based Research Agent Evaluation.}
Recent benchmarks evaluate deep research agents that perform multi-step information seeking and synthesize evidence-grounded reports~\citep{du2025deepresearch,bosse2025deepresearchbench,coelho2025deepresearchgym,han2025deer,reportbench}. They assess report quality---relevance, factuality, citation groundedness, coverage, and evidence use~\citep{du2025deepresearch,han2025deer,coelho2025deepresearchgym,wang2026deepresearcheval,sharma2025researchrubrics}---as well as process behavior, including search decisions, source selection, trajectory validity, and step-level reasoning~\citep{shao2025seekbench,ye2026miroevalbenchmarkingmultimodaldeep,chen2026srrjudge,zhan2026whyyour,hu2025stepdeepresearch,zhang2026carr,shao2025drtulu}. To scale beyond expert review, they increasingly rely on LLM judges for reports, citations, and evidence traces~\citep{zheng2023judgingNeurIPS,liu2023geval,dubois2023alpacafarm,coelho2025deepresearchgym,han2025deer,agentasjudge}. Our work complements this setting by meta-evaluating such judges under controlled process- and outcome-level failures.

\vspace{-0.5ex}
\textbf{Meta-Evaluation for LLM Judges.}
Another line of work meta-evaluates LLM judges and reward models using preference pairs, ranking tasks, verification settings, or trajectory-level annotations, including RewardBench2~\citep{malik2026rewardbench}, JudgeBench~\citep{tan2025judgebench}, JETTS~\citep{zhou2025evaluating}, VerifyBench~\citep{li2025verifybench}, AgentRewardBench~\citep{lu2025agentrewardbench}, and Sage~\citep{feng2025sage}. These benchmarks are informative but usually evaluate complete responses rather than localized failures in extended agent executions~\citep{tan2025judgebench,liu2025reife}. LLMBar~\citep{zeng2024llmbar} and ReIFE~\citep{liu2025reife} are closest, using clean--flawed adversarial pairs to isolate evaluation errors across models and protocols. However, they mainly target response-level instruction deviations, while research agents can fail during search, tool use, evidence selection, and synthesis~\citep{zhan2026whyyour,chen2026srrjudge,shao2025seekbench,lan2025deepwidesearch}. Our \benchmark{} investigates open-ended agent executions in non-verifiable settings.

%% file: sections/conclusion.tex
\section{Conclusion}
\label{sec:conclusion}
\vspace{-1ex}
We introduced \benchmark{}, a meta-evaluation benchmark for assessing whether
LLM judges can reliably evaluate evidence-based research agents. By constructing
verified reference executions and controlled failure-bearing alternatives,
\benchmark{} provides fine-grained labels over both process-level trajectory
 and outcome-level report failures. Our experiments show that current
judges remain limited across reasoning, tool use, and final-report evaluation.
They also reveal substantial variation across failure types, evaluation
granularity, prompting formats, and best-of-\(N\) selection settings. Overall,
these results suggest that judge reliability should be evaluated as a first-class
property of research-agent evaluation pipelines. Fine-grained protocols improve
failure sensitivity, but robust judge evaluation still requires careful choices
about model capability, cost, scoring interface, and evaluation unit. 
\vspace{-2ex}
\paragraph{Limitations.}
\benchmark{} is a meta-evaluation benchmark, so its scope is intentionally
controlled. The taxonomy covers common failures from current research-agent
traces and prior evaluation work, but it cannot exhaust every domain-specific, naturally occurring or
interactive failure mode. The controlled degradations isolate target failures to
make judge behavior measurable, complementing human audit studies of naturally
occurring agent errors. Like many modern benchmarks, \benchmark{} requires updating for longer term reliability: As judge models and research-agent systems evolve, the benchmark should be also updated with new traces, failure types, and evaluator families.

%% file: sections/contribution.tex
\section{Author Contributions}
\label{sec:author}
We summarize each author's primary contributions to the project below. Authors shown in \textbf{bold} took the lead role in the corresponding category.

\begin{itemize}[leftmargin=1pt, itemindent=*, labelsep=0.5em]
\item Project leadership: \textbf{Leyao Wang}, Yanan He
\item Core contributions: Leyao Wang, Yanan He, Peng Chen, Asaf Yehudai, Arman Cohan
\item \benchmark{} development (Process-Reasoning): Leyao Wang
\item \benchmark{} development (Process-Tool Use): Peng Chen
\item \benchmark{} development (Outcome): Yanan He
\item Evaluations and baselines: Leyao Wang, Yanan He, Peng Chen
\item Paper writing: Leyao Wang, Yanan He, Arman Cohan
\item Administration and policy review: Leyao Wang, Yixin Liu
\item Advising and mentorship: \textbf{Arman Cohan}, Asaf Yehudai, Yixin Liu, Rex Ying,  Michal Shmueli-Scheuer
\end{itemize}

Core contributors made sustained and significant contributions throughout the project.
All authors contributed to project discussions, experiment planning, and manuscript reviewing.

%% file: appendix/appendix.tex
\appendix

\section{Taxonomy Details and Related Work}
\label{app:taxonomy_details}

\subsection{Full Perturbation Taxonomy}
\label{app:full_taxonomy}

Table~\ref{tab:full_taxonomy} provides the complete definitions of all process-level and outcome-level perturbation types used in our benchmark.
 
\begin{table*}[h!]
\caption{Full definitions for all perturbation types in our taxonomy.}
\label{tab:full_taxonomy}
\centering
\small
\resizebox{\textwidth}{!}{%
\begin{tabular}{p{2.2cm}p{1.8cm}p{3.2cm}p{7.5cm}}
\toprule
\textbf{Module} & \textbf{Category} & \textbf{Error Type} & \textbf{Definition} \\
\midrule
\multicolumn{3}{@{}l}{\textit{Process-level errors}} \\
\midrule
Reasoning & Structure & Execution Stagnation & Consecutive search rounds repeat the same terms or fail to build on prior findings, causing the search process to loop without expanding coverage. \\
\addlinespace
Reasoning & Analysis & Shallow Reflection & Summarizes prior results without identifying knowledge gaps or adjusting the subsequent search direction, adding little analytical value. \\
\addlinespace
Reasoning & Faithfulness & Evidence Omission & Relevant evidence is available in the collected sources but is not incorporated into the final answer, resulting in incomplete synthesis. \\
\addlinespace
Reasoning & Faithfulness & Evidence Fabrication & Fabricates citations, findings, or author positions that are not present in any retrieved source. \\
\addlinespace
Tool & Structure & Wrong Tool Selection & The agent invokes a tool whose capability does not match the user's information need. \\
\addlinespace
Tool & Faithfulness & Constraint Drop & The tool call omits one or more constraints implied by the user's information need, causing the retrieved content to fall outside the user's requested scope. \\
\addlinespace
Tool & Faithfulness & Argument Corruption & A tool-call argument contains an incorrect value while preserving the argument structure. Includes named-entity errors and numeric or temporal values that deviate from the user's intended specification. \\
\addlinespace
Tool & Faithfulness & Result Irrelevance & The content returned by the tool falls outside the scope defined by the call's arguments. \\
\addlinespace
Tool & Groundedness & Wrong Source Citation & A claim in the response is attributed to a specific retrieved source, but the source's actual content does not support the paired claim. \\
\addlinespace
Tool & Groundedness & Tool Response Hallucination & The response contains a fact or entity-claim binding that is not grounded in any retrieved source. \\
\midrule
\multicolumn{3}{@{}l}{\textit{Outcome-level errors}} \\
\midrule
Output & Relevance & Incomplete Coverage
& The response does not adequately cover the key aspects of the user's question. Some aspects may be missing entirely, while others may be mentioned only briefly or without enough detail. \\
\addlinespace

Output & Relevance & Topical Misalignment
& The response includes content that is not directly relevant to the user's question, or gradually drifts away from the requested topic. \\
\addlinespace

Output & Faithfulness& Citation Groundedness
& The response uses a citation that is incorrect or unsupported, such as a fake citation, wrong citation number, misattributed source, or a citation that does not actually support the claim. \\
\addlinespace

Output & Faithfulness& Evidence Omission
& The response states a conclusion or important claim without providing sufficient evidence, examples, citations, or supporting details. \\
\addlinespace

Output & Faithfulness& Fabrication
& The response presents facts, findings, conclusions, examples, or relationships that are demonstrably false, invented, or attributed to the wrong entity/source. \\
\addlinespace

Output & Expression & Expression Quality
& The response has problems in readability, clarity, or language quality, such as awkward phrasing, repetition, or unnecessary verbosity. \\
\addlinespace

Output & Synthesis & Incoherence
& The response does not form a coherent whole due to contradictions, unclear transitions, reasoning gaps, or poor structural organization. \\

\bottomrule
\end{tabular}
}
\end{table*}

\subsection{Related Work Coverage}
\label{app:taxonomy_related_work}

Tables~\ref{tab:process_error_taxonomy_related_work} and~\ref{tab:outcome_error_taxonomy_related_work} summarize how prior work motivates and overlaps with our process-level and outcome-level error taxonomy.
\begin{table}[h!]
\centering
\small
\setlength{\tabcolsep}{3pt}
\renewcommand{\arraystretch}{1.12}
\caption{Related works for process-level error taxonomy.}
\label{tab:process_error_taxonomy_related_work}
\begin{tabularx}{\linewidth}{@{}p{0.15\linewidth}p{0.22\linewidth}X@{}}
\toprule
\textbf{Category} & \textbf{Error Type} & \textbf{Paper Mentioning the Error Types} \\
\midrule

Structure & Execution Stagnation
& SRR-Judge (Coverage \& Improvement Potential; Query Appropriateness; Logical Structure)~\citep{chen2026srrjudge}\\

\addlinespace[0.6em]

Analysis & Shallow Reflection
& AgentErrorTaxonomy (Over-simplification / Incomplete Summary)~\citep{zhu2025where};
SRR-Judge (Coverage \& Improvement Potential; Logical Structure; Clarity \& Conciseness)~\citep{chen2026srrjudge};
DeepWideSearch (lack of reflection)~\citep{lan2025deepwidesearch} \\

\addlinespace[0.6em]

Faithfulness& Evidence Omission
& AgentErrorTaxonomy (Memory: Over-simplification / Incomplete Summary)~\citep{zhu2025where};
DeepHalluBench (Hallucination)~\citep{zhan2026whyyour};
SeekBench (Groundedness)~\citep{shao2025seekbench} \\

\addlinespace[0.6em]

Faithfulness& Evidence Fabrication
& AgentErrorTaxonomy (Hallucination)~\citep{zhu2025where};
DeepHalluBench (Hallucination)~\citep{zhan2026whyyour};
SeekBench (Groundedness)~\citep{shao2025seekbench} \\

\addlinespace[0.6em]

Structure & Wrong Tool Selection
& AgentErrorTaxonomy (Unnecessary Tool; Missing Tool)~\citep{zhu2025where};
MetaTool (Tool Selection)~\citep{huang2024metatool};
ToolBeHonest (Tool-Selection Hallucination)~\citep{zhang2024toolbehonest};
BFCL (Function Selection; Relevance Detection)~\citep{patil2025bfcl} \\

\addlinespace[0.6em]

Faithfulness & Constraint Drop
& AgentErrorTaxonomy (Constraint Ignorance)~\citep{zhu2025where};
ComplexFuncBench (Implicit Parameter Reasoning)~\citep{zhong2025complexfuncbench};
SRR-Judge (Query Appropriateness)~\citep{chen2026srrjudge};
ToolSandbox (Insufficient Information)~\citep{lu2025toolsandbox} \\

\addlinespace[0.6em]

Faithfulness & Argument Corruption
& AgentErrorTaxonomy (Incorrect Argument)~\citep{zhu2025where};
SpecTool (Incorrect Argument Value; Name; Type)~\citep{kokane2024spectool};
BFCL (Parameter-Value Correctness)~\citep{patil2025bfcl};
ToolBeHonest (Tool Format Hallucination; Tool Content Hallucination)~\citep{zhang2024toolbehonest};
ToolSandbox (Time-related Argument Hallucinations; Named-Entity Errors)~\citep{lu2025toolsandbox} \\

\addlinespace[0.6em]

Faithfulness & Result Irrelevance
& RAGAs (Context Relevance)~\citep{es2023ragas};
ARES (Context Relevance)~\citep{saadfalcon2024ares};
RGB (Noise Robustness; Negative Rejection)~\citep{chen2023benchmarking};
SeekBench (Recovery from Low-Quality Evidence)~\citep{shao2025seekbench} \\

\addlinespace[0.6em]

Groundedness & Wrong Source Citation
& ALCE (Citation Precision; Citation Recall)~\citep{gao2023alce};
AttrScore (Attributable; Extrapolatory; Contradictory)~\citep{yue2024attrscore};
LongCite (Citation F1)~\citep{zhang2024longcite};
DeepResearch Bench (Citation Accuracy)~\citep{du2025deepresearch};
DEER (Cited Claim Verification)~\citep{han2025deer} \\

\addlinespace[0.6em]

Groundedness & Tool Response Hallucination
& RAGTruth (Baseless Info; Conflict-with-Context)~\citep{niu2024ragtruth};
FActScore (Atomic Fact Support)~\citep{min2023factscore};
DeepHalluBench (PIES Taxonomy)~\citep{zhan2026whyyour};
FaithEval (Contextual Faithfulness)~\citep{ming2025faitheval};
SAFE (Long-form Factuality)~\citep{wei2024longfact} \\

\bottomrule
\end{tabularx}
\end{table}

\begin{table}[t!]
\centering
\small
\setlength{\tabcolsep}{3pt}
\renewcommand{\arraystretch}{1.12}
\caption{Related works for outcome-level error taxonomy.}
\label{tab:outcome_error_taxonomy_related_work}
\begin{tabularx}{\linewidth}{@{}p{0.15\linewidth}p{0.22\linewidth}X@{}}
\toprule
\textbf{Category} & \textbf{Error Type} & \textbf{Paper Mentioning the Error Types} \\
\midrule

Relevance & Incomplete Coverage
& HaluQuestQA (Completeness; incomplete information)~\citep{haluQuestQA};
Expert Schema (Incomplete Answer; Major omissions; Lacking details)~\citep{martinboyle2026expert};
Dr. Bench (Information coverage; Informational coverage \& content depth)~\citep{yao2025rigorous};
ResearchRubrics (Completeness; rubric-item coverage)~\citep{sharma2025researchrubrics} \\

\addlinespace[0.6em]

Relevance & Topical Misalignment
& Dr. Bench (Topical Focus; SemanticDrift)~\citep{yao2025rigorous};
DRSE (Answer Relevance)~\citep{hwang2026shallow};
HaluQuestQA (Relevance)~\citep{haluQuestQA};
Expert Schema (Question redirection; Question misinterpretation)~\citep{martinboyle2026expert} \\

\addlinespace[0.6em]

Faithfulness& Citation Groundedness
& DRSE (Citation Precision; Citation Recall)~\citep{hwang2026shallow};
DeepResearch Bench (Effective Citation Count; Overall Citation Accuracy)~\citep{du2025deepresearch};
Dr. Bench (Retrieval Trustworthiness; Trustworthy-Source Links)~\citep{yao2025rigorous};
Expert Schema (Citation information; Source confusion; Incomplete references; Inconsistent referencing)~\citep{martinboyle2026expert} \\

\addlinespace[0.6em]

Faithfulness& Evidence Omission
& HaluQuestQA (Completeness; References; incomplete information)~\citep{haluQuestQA};
Expert Schema (Incomplete references; Lacking details; Incomplete Answer)~\citep{martinboyle2026expert};
Dr. Bench (Citation quality \& source credibility; source verification; evidence organization)~\citep{yao2025rigorous};
DEER (Evidence Coverage; Information Sufficiency)~\citep{han2025deer} \\

\addlinespace[0.6em]

Faithfulness& Fabrication
& Expert Schema (Contains hallucinations; Basic accuracy issues)~\citep{martinboyle2026expert};
Dr. Bench (factual accuracy; source verification)~\citep{yao2025rigorous};
HaluQuestQA (Factuality; factual inconsistencies)~\citep{haluQuestQA}  \\

\addlinespace[0.6em]

Expression & Expression Quality
& Expert Schema (Verbosity; Language issues; Notation errors)~\citep{martinboyle2026expert};
Dr. Bench (logical clarity \& expression; formatting consistency)~\citep{yao2025rigorous};
ResearchRubrics (Clarity)~\citep{sharma2025researchrubrics} \\

\addlinespace[0.6em]

Synthesis & Incoherence
& Expert Schema (Self-contradiction; Disjointed response)~\citep{martinboyle2026expert};
Dr. Bench (structural organization; information integration)~\citep{yao2025rigorous};
DRSE (Organization)~\citep{hwang2026shallow};
ResearchRubrics (cross-document synthesis; reasoning soundness; clarity)~\citep{sharma2025researchrubrics} \\

\bottomrule
\end{tabularx}
\end{table}
\section{Benchmark Construction and Validation}
\label{app:benchmark_construction}

This section provides additional details on how benchmark instances are organized and validated. 
We report the distribution of perturbation types to make clear how many paired examples are available for each failure mode, and we describe the validation process used to ensure that each perturbed instance reflects the intended error type.

\subsection{Dataset Statistics}
\label{app:dataset_statistics}

Table~\ref{tab:dataset_statistics} reports the number of perturbation pairs for each failure type in our benchmark. 
The distribution is approximately balanced across the process-level (reasoning and tool-use) and outcome-level error types.

\begin{table}[t!]
\centering
\caption{Dataset statistics for \benchmark perturbation types.}
\label{tab:dataset_statistics}
\footnotesize
\begin{tabular}{lllrr}
\toprule
\textbf{Target} & \textbf{Category} & \textbf{Perturbation Type} &
\textbf{\# Pairs} & \textbf{\%} \\
\midrule
\multicolumn{5}{l}{\textit{Reasoning process perturbations}} \\
Reasoning & Faithfulness & Evidence Fabrication & 35 & 25.00 \\
Reasoning & Faithfulness & Evidence Omission    & 36 & 25.71 \\
Reasoning & Analysis     & Shallow Reflection   & 40 & 28.57 \\
Reasoning & Structure    & Execution Stagnation & 29 & 20.71 \\
\cmidrule(lr){4-5}
Reasoning & \textbf{Total} & -- & \textbf{140} & \textbf{100.00} \\
\midrule
\multicolumn{5}{l}{\textit{Tool-use perturbations}} \\
Tool Use & Structure    & Wrong Tool Selection          & 11 & 8.33 \\
Tool Use & Faithfulness & Constraint Drop               & 11 & 8.33 \\
Tool Use & Faithfulness & Argument Corruption           & 28 & 21.21 \\
Tool Use & Faithfulness & Result Irrelevance            & 28 & 21.21 \\
Tool Use & Groundedness & Wrong Source Citation         & 28 & 21.21 \\
Tool Use & Groundedness & Tool Response Hallucination   & 26 & 19.70 \\
\cmidrule(lr){4-5}
Tool Use & \textbf{Total} & -- & \textbf{132} & \textbf{100.00} \\
\midrule
\multicolumn{5}{l}{\textit{Outcome-level perturbations}} \\
Outcome & Faithfulness& Citation Groundedness & 26 & 13.00 \\
Outcome & Faithfulness& Evidence Omission     & 28 & 14.00 \\
Outcome & Faithfulness& Fabrication           & 29 & 14.50 \\
Outcome & Expression & Expression Quality    & 30 & 15.00 \\
Outcome & Relevance  & Incomplete Coverage   & 28 & 14.00 \\
Outcome & Relevance  & Topical Misalignment  & 30 & 15.00 \\
Outcome & Synthesis  & Incoherence           & 29 & 14.50 \\
\cmidrule(lr){4-5}
Outcome & \textbf{Total} & -- & \textbf{200} & \textbf{100.00} \\
\bottomrule
\end{tabular}

\end{table}

\subsection{Human-in-the-Loop Calibration and Curation}
\label{app:perturbation_validation}

We validate perturbations to ensure that each edited instance introduces the intended failure type while preserving the surrounding execution or report context. 
This validation step checks that the reference instance does not already contain the target failure at the selected edit site, and that the perturbed instance reflects the controlled degradation rather than an unrelated change. 
The following examples illustrate representative perturbations and validation cases across our taxonomy.

\begin{figure}[h!]
    \centering
    \includegraphics[width=1\linewidth]{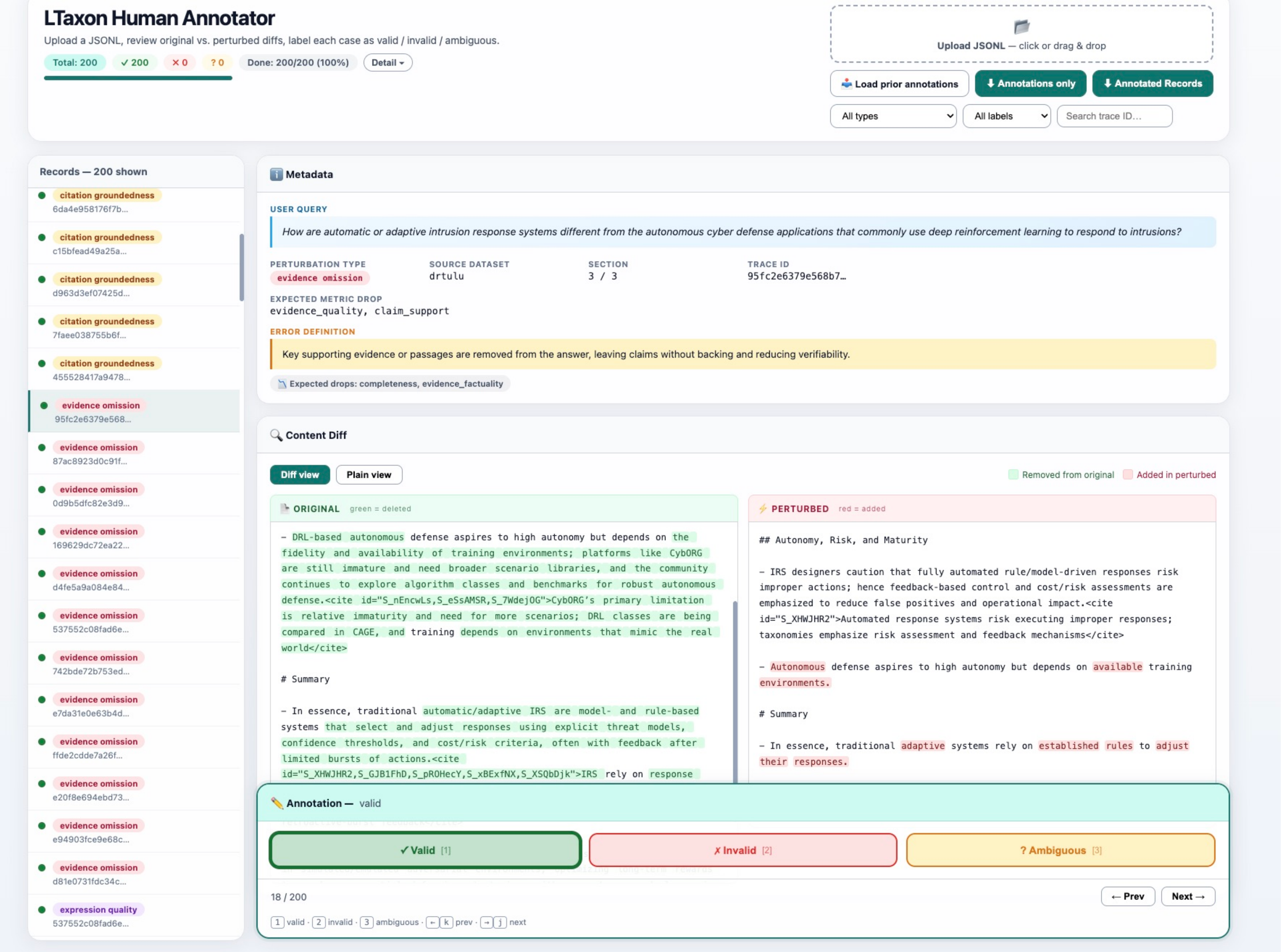}
    \caption{
    Human annotation interface for perturbation validation. Annotators review the user query, target perturbation type, error definition, expected metric drops, and side-by-side diff between the original and perturbed content. They then label each perturbation as valid, invalid, or ambiguous.
    }
    \label{fig:human_annotation_interface}
\end{figure}

\paragraph{Human Annotation Interface.}
Figure~\ref{fig:human_annotation_interface} shows the annotation interface used in our validation process. The top panel displays metadata for the instance, including the user query, source dataset, trace identifier, perturbation type, and expected metric drops. The middle panel shows a side-by-side comparison of the original and perturbed text, with deleted spans highlighted in green and inserted spans highlighted in red. The bottom panel allows annotators to assign one of three labels: valid, invalid, or ambiguous. This design helps annotators judge whether the perturbation is aligned with the target error definition and whether the local edit produces the intended degradation in relevance, factuality, coherence, coverage, or expression quality.

\newpage
\section{Implementation Details and Prompts}
\label{app:implementation_details}







\subsection{Pointwise Judge Prompts}
\label{app:process_pointwise_judge_prompts}

\FloatBarrier

\input{appendix/prompts}

\newpage
\section{Taxonomy Validation: Case Studies and Perturbation Examples}
\label{app:case_study}
\input{appendix/Error_type_case}
\input{appendix/process_natual_cases}











\section{Ethics and Broader Impact}
\label{sec:ethics_broader_impact}

This work studies the reliability of LLM judges for evidence-based deep research agents. As such systems are increasingly used to support information seeking, report generation, and research-oriented workflows, reliable evaluation is important for both scientific progress and responsible deployment. A potential positive impact of REFLECT is that it provides a more diagnostic way to evaluate judge models: rather than relying only on coarse human-preference agreement or aggregate scores, it tests whether judges can detect localized failures in reasoning, tool use, grounding, factuality, and synthesis. This may help researchers and practitioners identify evaluator blind spots, design more robust evaluation protocols, and avoid overestimating the trustworthiness of automated research agents.

At the same time, REFLECT is a meta-evaluation benchmark rather than a guarantee of judge reliability in all real-world settings. Strong performance on controlled perturbations should not be interpreted as sufficient evidence for safe use in high-stakes domains such as medicine, law, finance, public policy, or scientific decision-making, where automated judge outputs should be combined with human oversight, domain-expert review, and additional evaluations on naturally occurring errors. Detailed failure taxonomies and perturbation examples may also create risks of benchmark gaming or overfitting to known error patterns, and the underlying traces, reports, model outputs, and annotations may reflect biases or coverage gaps from the source data and model families used in the study. We therefore encourage future work to expand and update the taxonomy, document data and model provenance carefully, and study fairness, robustness, privacy, and safety implications in more detail.

%% file: appendix/prompts.tex
\begin{tcolorbox}[
  promptbox,
  title={HOLISTIC\_RUBRIC\_NO\_COT\_PROMPT}
]
\small
\begin{verbatim}
You are an expert evaluator of AI research agent traces.

Given a research query and the agent's full reasoning trace (including
planning, search, reflection, and synthesis steps), rate the trace on the
following dimensions. Use a 1-5 scale where 1=very poor and 5=excellent.

Dimensions:
- coherence: Does the reasoning flow logically?
  Do later steps build on earlier findings?
- factual_alignment: Are claims accurate and consistent with the retrieved
  evidence?
- info_density: Does the trace contain specific facts, numbers, and concrete
  details?
- synthesis_quality: Does the final answer integrate information across sources
  with comparative analysis?

Query: {query}

Agent Trace:
{trace}

Output ONLY a JSON object with four integer scores, no other text:
{{"coherence": ..., "factual_alignment": ..., "info_density": ...,
  "synthesis_quality": ...}}
\end{verbatim}
\end{tcolorbox}

\begin{tcolorbox}[
  promptbox,
  title={FINE-GRAINED\_RUBRIC\_NO\_COT\_PROMPT}
]
\small
\begin{verbatim}
You are an expert evaluator of AI research agent reasoning steps.

You are given a research query, the prefix of an agent's reasoning trace
(all steps completed so far), and a single candidate next step. Your task is
to evaluate how good that candidate step is as the immediate next step.

Rate the candidate step on three dimensions using a 1-5 scale
(1=very poor, 5=excellent):
- relevance: Does the step follow logically from the prefix and remain directly
  on-topic for the query?
- quality: Is the step well-executed for its type? (e.g., a search step issues
  a precise, focused query; a planning step is clearly structured; a synthesis
  step integrates evidence coherently)
- contribution: Does the step add meaningful new information or meaningfully
  advance the reasoning toward answering the query?

Query: {query}

=== Trace Prefix (completed steps) ===
{prefix}

=== Candidate Next Step ===
{step}

Output ONLY a JSON object with three integer scores, no other text:
{{"relevance": ..., "quality": ..., "contribution": ...}}
\end{verbatim}
\end{tcolorbox}

\begin{tcolorbox}[
  promptbox,
  title={HOLISTIC\_RUBRIC\_NO\_RUBRIC\_COT\_PROMPT}
]
\small
\begin{verbatim}
You are an expert evaluator of AI research agent traces.

Given a research query and the agent's full reasoning trace, assess the
overall quality holistically. Use a 1-10 scale where 1=very poor and
10=excellent.

Consider all aspects of quality together: logical coherence, factual accuracy,
information richness, and how well the final answer synthesizes the evidence.

Query: {query}

Agent Trace:
{trace}

First, briefly analyze the trace's strengths and weaknesses in 2-3 sentences.
Then output your final score as a JSON object with a single integer score:
{{"overall": ...}}

Analysis:
\end{verbatim}
\end{tcolorbox}

\begin{tcolorbox}[
  promptbox,
  title={HOLISTIC\_RUBRIC\_NO\_RUBRIC\_NO\_COT\_PROMPT}
]
\small
\begin{verbatim}
You are an expert evaluator of AI research agent traces.

Given a research query and the agent's full reasoning trace (including
planning, search, reflection, and synthesis steps), rate the overall quality
of the trace holistically. Use a 1-10 scale where 1=very poor and 10=excellent.

Consider all aspects of quality together: logical coherence, factual accuracy,
information richness, and how well the final answer synthesizes the evidence.

Query: {query}

Agent Trace:
{trace}

Output ONLY a JSON object with a single integer score, no other text:
{{"overall": ...}}
\end{verbatim}
\end{tcolorbox}

\begin{tcolorbox}[
  promptbox,
  title={FINE-GRAINED\_NO\_RUBRIC\_NO\_COT\_PROMPT}
]
\small
\begin{verbatim}
You are an expert evaluator of AI research agent reasoning steps.

You are given a research query, the prefix of an agent's reasoning trace
(all steps completed so far), and a single candidate next step. Rate the
overall quality of that candidate step holistically. Use a 1-10 scale where
1=very poor and 10=excellent.

Consider all aspects together: whether the step follows logically from the
prefix, how well it is executed for its type, and how meaningfully it advances
reasoning toward the answer.

Query: {query}

=== Trace Prefix (completed steps) ===
{prefix}

=== Candidate Next Step ===
{step}

Output ONLY a JSON object with a single integer score, no other text:
{{"overall": ...}}
\end{verbatim}
\end{tcolorbox}

%% file: appendix/Error_type_case.tex
\subsection{Outcome-level Analysis}
\subsubsection{Error Case Studies}
\label{sec:output_error_case_studies}

The following cases are drawn from Deep Research Bench~\citep{du2025deepresearch}
and Tongyi DeepResearch~\citep{team2025tongyi}. Each example includes the
user query, selected response excerpts, and annotations based on our output-level
error taxonomy.

\noindent\textbf{Case 1: Citation Groundedness and Evidence Omission.}

\begin{tcolorbox}[inputbox]
Please prepare a market research analysis of the global video editing and
creation software/tool market. Include major products like those from Adobe
(Premiere Pro, After Effects), CapCut, DaVinci Resolve, Final Cut Pro, and
others relevant in the current landscape.
\end{tcolorbox}

\begin{tcolorbox}[outputbox, title={Model Response [Excerpts]}]
\small
\textit{[Excerpt --- Major Players and Market Shares]}

\smallskip
Adobe Premiere Pro: $\sim$35\% market share; controls the professional/enterprise
segment.
Final Cut Pro X: $\sim$25\% market share; preferred by professional Mac users.
DaVinci Resolve: $\sim$15\% market share; rapidly growing due to its free tier
and professional features.~[1][2][9][10][11]

\medskip
\textit{[Excerpt --- Market Size and Growth Trends]}

\smallskip
As of 2025, the global video editing software market is estimated between \$2.5
and \$3.65 billion, depending on scope, with projected CAGR between 5.2\% and
8.1\% through 2029. By 2029, forecasts place the market size between \$3.3
billion and \$5 billion, depending on whether services and AI tools are included
in the estimate.~[1][2][3][4][5] The AI video editing sector is the
fastest-growing segment, forecast to skyrocket from \$1.6 billion in 2025 to
\$9.3 billion by 2030 at over 42\% CAGR.~[6] Paid video editing software users
are projected to reach 48.2 million in 2025.~[7] Cloud-based editing solutions
are outpacing traditional models, with future CAGR outstripping on-premise
solutions (8.5\% for cloud-based platforms).~[1][2][8] Asia-Pacific is the
fastest-growing regional market (CAGR 7.5\%), while North America remains the
largest by market share.~[1][2][3][5]
\vspace{4pt}\noindent\rule{\linewidth}{0.3pt}\vspace{2pt}

{\footnotesize\RaggedRight
\noindent[1]~The Business Research Company. \textit{Global Video Editing Software Market Overview 2025.}

\noindent[2]~Mordor Intelligence. \textit{Video Editing Market Size, Share and Growth Research.}

\noindent[3]~Straits Research. \textit{Video Editing Software Market Size, Share \& Growth.}

\noindent[4]~DataIntelo. \textit{Video Editing Service Market Report.}

\noindent[5]~The Business Research Company. \textit{Audio And Video Editing Software Market 2025.}

\noindent[6]~Virtue Market Research. \textit{AI Video Editing Tools Market | Size, Share, Growth | 2025--2030.}

\noindent[7]~Triple A Review. \textit{Video Editing Statistics You Need to Know in 2025.}

\noindent[8]~SendShort. \textit{Video Editing Software Market Statistics (2025).}

\noindent[9]~PCMag. \textit{The Best Video Editing Software We've Tested.}

\noindent[10]~DIY Video Editor. \textit{Best Video Editing Software 2025 Reviewed and Compared.}

\noindent[11]~Project Aeon. \textit{Best AI-Video Maker: Top Tools for 2025.}

\vspace{2pt}
\noindent\textit{For readability and clarity, we omit the source URLs and only show the cited source content.}
}
\end{tcolorbox}

\noindent\textbf{\textsc{Citation Groundedness}}: The response contains many
citations, but several precise market claims are not clearly grounded in strong
or directly relevant sources. For example, it gives exact-looking market shares
such as Adobe Premiere Pro $\sim$35\%, Final Cut Pro X $\sim$25\%, and DaVinci
Resolve $\sim$15\%, but the surrounding sources are mostly market overview pages,
software review articles, or general industry reports, not clearly authoritative
market-share evidence for those exact percentages. It also cites weaker sources
such as LinkedIn posts, blog-style articles, Reddit, and product-review pages for
market-level claims. This creates a gap between the specificity of the claims and
the quality/directness of the citations.

\medskip
\noindent\textbf{\textsc{Evidence Omission}}: The report makes many quantitative
claims, such as the global market being \$2.5--\$3.65 billion, AI video editing
growing from \$1.6B to \$9.3B, paid users reaching 48.2 million, and
Asia-Pacific having 7.5\% CAGR. These claims are presented as facts, but the
response does not explain how the estimates were derived, whether different
reports define the market differently, or why the ranges vary so much. The answer
gives numbers, but does not provide enough methodological context or evidence
detail to support them.

\noindent\textbf{Case 2: Expression Quality, Incoherence, and Topical Misalignment.}

\begin{tcolorbox}[inputbox]
What is the position of Canada when it comes to Lethal Autonomous Weapon Systems,
from both a moral and strategic perspective and what regulations does it impose?
\end{tcolorbox}

\begin{tcolorbox}[outputbox, title={Model Response [Excerpts]}]
\small
\textit{[Excerpt 1 --- Ethical and Moral Perspectives (early section, coherent)]}

\smallskip
Canada asserts that developments in military AI and autonomy are ethically bounded by a
core premise: decisions on whether to use lethal force must always rest with human
authorities. Its government leadership acknowledges grave risks posed by machines that
operate without direct or effective human input---particularly to civilian populations and
fundamental notions of human dignity. References to international standards such as the
Convention on Certain Conventional Weapons (CCW) inform this positioning, emphasizing
that only weapons subject to unambiguous ``appropriate human control'' will be compliant
with Canadian ethical priorities and humanitarian obligations.

\medskip
\textit{[Excerpt 2 --- Governance Structure Overview (later section, degraded)]}

\smallskip
Canada institutionalizes its commitment through specialized boards overseeing cyber-risks
identified via periodic exercises assessing systemic vulnerabilities inherent within
interconnected network segments supporting critical assets needing constant protective
layer reinforcement regardless status assigned classified as low medium high sensitivity
affecting overall survivability performance levels exhibited sustained throughout extended
operational periods subject always careful risk/benefit assessments executed centrally via
top-down mandate directives transmitted downward reflecting bottom-up recommendations
derived accordingly depending on specific needs expressed clearly by end users constant
constant firm loyal devoted unwavering resolute determined courageous brave true valiant
heroic illustrious famous distinguished noteworthy honorable noble respectable virtuous
ethical moral upright honest sincere genuine trustworthy reliable dependable secure safe
protected free liberated liberated free liberated sovereign independent autonomous
autonomous autonomous autonomous liberated liberated colonization colonization colonization
colonization colonization colonization colonisation colonizn colonisation settlements
settlements bases bases outposts visualization visualization visualization conceptual
notion sense perception sensory sensory perception concept.
\end{tcolorbox}

\noindent\textbf{\textsc{Expression Quality}}: The response exhibits severe readability
degradation in its later sections. What begins as a structured policy discussion
deteriorates into multi-clause sentences with no discernible logical endpoint, such as the
``Governance Structure Overview'' passage, which accumulates subordinate clauses for over
200 words without completing a coherent thought. Further on, the text degenerates into
repetitive word strings (``autonomous autonomous autonomous autonomous,''
``liberated liberated liberated'') and taxonomically unrelated noun lists
(``colonization\ldots{}visualization\ldots{}communities''). These are not merely
stylistic imperfections; they render substantial portions of the response unreadable and
unprofessional.

\medskip
\noindent\textbf{\textsc{Incoherence}}: The response does not form a coherent whole. It begins with a clear structure, discussing Canada’s ethical stance, support for restrictions on LAWS, and strategic safeguards. However, the second half abandons this framework and shifts into content with no clear argumentative purpose or organizational logic. The transitions become especially weak, moving through synonym-like phrases and loosely associated nouns rather than developing the original analysis. Although the conclusion tries to return to a coherent “dual commitment” narrative, the earlier breakdown in structure makes the overall response feel fragmented and poorly organized.

\medskip
\noindent\textbf{\textsc{Topical Misalignment}}: The user asks about Canada’s moral stance, strategic perspective, and regulations on LAWS. While the response starts on-topic, later sections drift into unrelated content such as natural disasters, supply chains, atmospheric phenomena, and programming languages, which have no connection to LAWS policy and fall outside the requested scope.

\noindent\textbf{Case 3: Incomplete Coverage and Incoherence.}

\begin{tcolorbox}[inputbox]
Research and analyze the diverse paths taken by various countries in Europe,
Asia, and the Americas to transition into the ranks of `developed nations'
following World War~II.
\end{tcolorbox}
\begin{tcolorbox}[outputbox, title={Model Response [Excerpts]}]
\small
\textit{[Excerpt --- Introduction]}

\smallskip
This analysis examines the trajectories of representative nations---West Germany
(Europe) and Japan (Asia)---emphasizing their starting points after WWII,
resource endowments, development strategies, and the influence of external and
internal factors.

\medskip
\textit{[Excerpt --- Patterns from the Americas]}

\smallskip
While this analysis has focused on Germany and Japan, it is notable that postwar
economic hegemons in the Americas, particularly the United States and Canada,
shared several but not all of these characteristics: The U.S.\ and Canada entered
the postwar period with intact infrastructure, booming demand, and a head start
as principal victors---and so their challenge was less about reconstruction and
more about managing growth, employment, and social inclusion.
\end{tcolorbox}

\noindent\textbf{\textsc{Incomplete Coverage}}: The response focuses almost
entirely on West Germany and Japan, while the prompt asks for diverse development
paths across Europe, Asia, and the Americas. Although Germany and Japan are
analyzed in depth, the Americas receive only a brief contextual mention, with no
substantive case study of the U.S., Canada, or any Latin American country.
Therefore, the answer does not fully cover the regional breadth requested by the
user.

\medskip
\noindent\textbf{\textsc{Incoherence}}: The response does not form a fully
coherent or well-organized whole because its overall structure does not align with
the user's requested dimensions. The prompt asks for a comparative analysis across
Europe, Asia, and the Americas, but the answer is organized primarily around two
detailed case studies: West Germany and Japan. The Americas are only added as a
brief supplementary note rather than being integrated as a parallel regional
dimension. As a result, the response's structure does not coherently support the
requested multi-region comparison, even though the individual sections are locally
organized.

\noindent\textbf{Case 4: Fabrication.}

\begin{tcolorbox}[inputbox]
Does p53 regulate myc in \textit{Drosophila melanogaster}?
\end{tcolorbox}

\begin{tcolorbox}[outputbox, title={Model Response [Excerpt]}]
\small
\textit{[Excerpt --- Feedback Loops via MicroRNAs]}

\smallskip
Evidence exists that dMyc regulates miRNAs (such as miR-308 and miR-305
families), forming negative feedback loops involving dp53.
MiR-308 targets dp53 directly, allowing cells with high dMyc and low
dp53 to promote competitive advantage. However, this again emphasizes how Myc
controls dp53 instead of vice versa.
\end{tcolorbox}

\noindent\textbf{\textsc{Fabrication}}: It claims that miR-308 directly targets
dp53, but the known relationship is that miR-308 is associated with dMyc
regulation, while miR-305 is the miRNA linked to modulation of Dmp53/p53 levels.
This is not merely under-supported; it assigns the regulatory role to the wrong
miRNA and therefore creates a false mechanistic claim.

\subsubsection{Perturbation Examples}
\label{sec:output_level_perturbation_examples}
\providecommand{\orig}[1]{\textcolor{blue!70!black}{#1}}
\providecommand{\pert}[1]{\textcolor{red!75!black}{#1}}

To illustrate how the output-level taxonomy is operationalized during
perturbation construction, we provide representative examples of original and
perturbed answer spans. Blue text marks the original span targeted by the
perturbation, while red text marks the inserted or modified perturbed span. The
examples are designed to preserve surface fluency while inducing a controlled
degradation aligned with a specific output-level error type.

\noindent\textbf{Sample 1: Incomplete Coverage.}

\noindent\textbf{Query:} In FinTech, how do machine learning algorithms apply to asset allocation?

\smallskip
\noindent\textbf{Original:}
The core innovation of the Black-Litterman model lies in its application of
Bayesian statistics to estimate expected asset returns. Instead of relying solely
on historical data or requiring the user to specify a complete vector of expected
returns for all assets, as in MVO, the BL model starts with a
\orig{prior distribution for expected returns derived from market equilibrium.
This prior represents a neutral, objective starting point. This market-implied
prior is then updated with the investor's subjective views, which act as the
``data'' or ``observations'' in the Bayesian sense, to arrive at a posterior
distribution of expected returns.}

\smallskip
\noindent\textbf{Perturbed:}
The core innovation of the Black-Litterman model lies in its application of
Bayesian statistics to estimate expected asset returns. This approach
systematically combines \pert{a market-implied prior distribution of expected
returns with an investor's subjective views to produce a refined posterior
distribution.}

\smallskip
\noindent\textbf{\textsc{Perturbation Rationale}:}
The perturbation compresses the mechanism into a high-level summary. It preserves
the surface topic of the Black-Litterman model, but removes the explanation of
why market equilibrium defines the prior and how investor views function as
Bayesian observations. As a result, the answer remains structurally relevant but
loses substantive coverage of how the method works.

\medskip

\noindent\textbf{Sample 2: Topical Misalignment.}

\noindent\textbf{Query:} Institutional Drivers of Digital Integration into Ethiopian Higher Education

\smallskip
\noindent\textbf{Original:}
The pandemic demonstrated that one-off directives cannot substitute for sustained
\orig{institutional readiness; where infrastructure is thin and users have limited
digital literacy, transitions to blended or online modalities are fragile and
inequitable.}

\smallskip
\noindent\textbf{Perturbed:}
The pandemic demonstrated that emergency remote teaching demands resilient
\pert{techno-pedagogical scaffolding; when Learning Management Systems lack
optimized asynchronous architectures, cognitive load overwhelms students
navigating poorly integrated multimedia content.}

\smallskip
\noindent\textbf{\textsc{Perturbation Rationale}:}
The perturbation preserves the broad pandemic-and-digital-education context, but
shifts the focus from institutional readiness and equity to LMS design and
cognitive load, creating a subtle topical drift away from the requested
institutional drivers.

\medskip

\noindent\textbf{Sample 3: Citation Groundedness.}

\noindent\textbf{Query:} In FinTech, how do ML algorithms apply to asset allocation? (Black-Litterman model)

\smallskip
\noindent\textbf{Original:}
The BL model assumes that the market portfolio is optimal. \orig{[6]} Using
reverse MVO, the model calculates the implied equilibrium excess returns given
the market's covariance matrix and risk aversion. \orig{[7]}

\smallskip
\noindent\textbf{Perturbed:}
The BL model assumes that the market portfolio is optimal. \pert{[7]} Using
reverse MVO, the model calculates the implied equilibrium excess returns given
the market's covariance matrix and risk aversion. \pert{[6]}

\smallskip
\noindent\textbf{\textsc{Perturbation Rationale}:}
The perturbation swaps the inline citation markers [6] and [7], breaking the connection between the claim and its
supporting source.

\medskip

\noindent\textbf{Sample 4: Evidence Omission.}

\noindent\textbf{Query:} Link SDG6 target 6.a with other SDG targets

\smallskip
\noindent\textbf{Original:}
\ldots{}from health, education, and gender \orig{(SDG 3, 4, 5)} to agriculture
and energy \orig{(SDG 2, 7)}, resilient cities and climate \orig{(SDG 11, 13)},
pollution control and biodiversity \orig{(SDG 12, 14, 15)}, and inclusive
growth and innovation \orig{(SDG 8, 9)}. \orig{Closing the financing and
capacity gaps identified in the SDG 6 Synthesis Report is thus a practical
pathway for using 6.a to accelerate the wider 2030 Agenda.}

\smallskip
\noindent\textbf{Perturbed:}
\ldots{}which UN syntheses describe as central to progress across all SDGs.
\pert{The research literature converges on this broad interconnectedness, though
the precise inter-goal mechanisms continue to be refined as new evidence
emerges.}

\smallskip
\noindent\textbf{\textsc{Perturbation Rationale}:}
The perturbation preserves the general claim that SDG 6.a is connected to other
SDGs, but removes the specific cross-SDG mappings and synthesis-report support
that substantiate the claim.

\medskip

\noindent\textbf{Sample 5: Fabrication.}

\noindent\textbf{Query:} Are there differences in uptake and benefits of a
Selenium-rich diet of biofortified crops vs. inorganic Se salt supplementation?

\smallskip
\noindent\textbf{Original:}
Agronomic evidence shows that foliar \orig{selenate} fertilization efficiently
supports selenium biofortification in crops.

\smallskip
\noindent\textbf{Perturbed:}
Agronomic evidence shows that foliar \pert{selenite} fertilization efficiently
supports selenium biofortification in crops.

\smallskip
\noindent\textbf{\textsc{Perturbation Rationale}:}
The perturbation swaps one selenium compound for another mechanistically
different compound, creating a domain-specific factual error.

\medskip
\noindent\textbf{Sample 6: Expression Quality.}

\noindent\textbf{Query:} Explain why a first-order ODE system may have infinitely many solutions.

\smallskip
\noindent\textbf{Original:}
A system of first-order ODEs, like the one derived in Section 5.3,
generally admits an infinite number of solutions.

\smallskip
\noindent\textbf{Perturbed:}
A system of first-order ODEs, like the one derived in Section 5.3,
generally \pert{admit} an infinite number of solutions.

\smallskip
\noindent\textbf{\textsc{Perturbation Rationale}:}
The perturbation introduces a subject--verb agreement error by replacing
``admits'' with ``admit.'' Since the grammatical subject is the singular noun
phrase ``A system,'' the verb should also be singular. This change does not alter
the underlying mathematical claim, but it reduces grammatical correctness,
fluency, and professional writing quality.

\medskip

\noindent\textbf{Sample 7: Incoherence.}

\noindent\textbf{Query:} vehicle routing algorithm supply and demand considering congestion

\smallskip
\noindent\textbf{Original:}
Choosing the congestion model: use \orig{time-dependent travel times when
congestion is primarily exogenous/predictable}, and \orig{flow-dependent travel
times with equilibrium when the fleet's routing materially affects traffic}.

\smallskip
\noindent\textbf{Perturbed:}
Choosing the congestion model: use \pert{flow-dependent travel times with
equilibrium when congestion is primarily exogenous/predictable}, and
\pert{time-dependent travel times when the fleet's routing materially affects
traffic}.

\smallskip
\noindent\textbf{\textsc{Perturbation Rationale}:}
The perturbation reverses the mapping between congestion conditions and modeling
choices. As a result, each congestion scenario is paired with the modeling choice
intended for the opposite case, creating an internally incoherent recommendation.

%% file: appendix/process_natual_cases.tex
\subsection{Process-Level Analysis}
\subsubsection{ Error Case Studies}
\label{sec:process_error_case_studies}

The following cases are drawn from rollout traces for the query
``Does p53 regulate myc in \textit{Drosophila melanogaster}?'' Each example
includes selected process excerpts and annotations based on our process-level
error taxonomy.

\noindent\textbf{Case 1: Execution Stagnation.}

\begin{tcolorbox}[inputbox]
Does p53 regulate myc in \textit{Drosophila melanogaster}?
\end{tcolorbox}

\begin{tcolorbox}[outputbox, title={Rollout Process [Excerpt]}]
\small
\textit{[Excerpt --- Repeated retrieval attempts around the same source]}

\smallskip
The rollout repeatedly searches for and attempts to access the same paper and
closely related variants of the same evidence target, including queries such as
``Supercompetitor status of Drosophila Myc cells requires p53 abstract,''
followed by repeated visits to PubMed, PMC, Cell, ScienceDirect, Jina proxy, and
NCBI Eutils pages for the same or near-identical article.

\smallskip
Many of these visits return inaccessible-page failures such as
``webpage content could not be accessed.'' Despite these failures, the rollout
continues to issue similar search and visit calls around the same article rather
than substantially changing the search strategy or broadening the evidence base.
\end{tcolorbox}

\noindent\textbf{\textsc{Execution Stagnation}}: The search process loops around
the same evidence target after repeated access failures. Rather than using the
failed visits as a signal to reformulate the search direction, seek alternative
review articles, inspect different experimental contexts, or explicitly separate
direct regulation from indirect genetic interaction, the rollout keeps trying
near-duplicate queries and access paths for the same paper. This matches the
definition of execution stagnation: consecutive retrieval rounds repeat similar
terms and fail to build on prior findings, causing the search process to expend
many steps without meaningfully expanding coverage.

\medskip

\noindent\textbf{Case 2: Shallow Reflection.}

\begin{tcolorbox}[inputbox]
Does p53 regulate myc in \textit{Drosophila melanogaster}?
\end{tcolorbox}

\begin{tcolorbox}[outputbox, title={Rollout Process [Excerpt]}]
\small
\textit{[Excerpt --- Reflection after repeated access failures]}

\smallskip
After many failed page visits, the rollout reflects that the relevant pages
cannot be accessed and states, in effect, that it must proceed using search
snippets. The process then continues toward a full mechanistic answer rather than
explicitly identifying unresolved evidence gaps or adjusting the search plan.

\smallskip
For example, the rollout acknowledges that full-text access has failed, but then
continues to synthesize claims about p53--dMyc regulation, apoptotic pathways,
and feedback mechanisms using partial snippets and background assumptions.
\end{tcolorbox}

\noindent\textbf{\textsc{Shallow Reflection}}: The reflection identifies a
surface-level retrieval problem, but it does not translate that observation into
a stronger reasoning adjustment. A deeper reflection would distinguish which
subquestions remain unresolved, such as whether dp53 directly regulates
\textit{dMyc} transcription, whether dMyc regulates dp53, whether the observed
relationship is limited to apoptosis or cell competition, and whether available
evidence supports direct regulation or only indirect interaction. Instead, the
rollout merely notes the access limitation and proceeds with a confident final
synthesis. This adds little analytical value beyond summarizing that retrieval
failed.

\medskip

\noindent\textbf{Case 3: Evidence Omission.}

\begin{tcolorbox}[inputbox]
Does p53 regulate myc in \textit{Drosophila melanogaster}?
\end{tcolorbox}

\begin{tcolorbox}[outputbox, title={Rollout Evidence [Excerpt]}]
\small
\textit{[Excerpt --- Retrieved evidence not fully incorporated]}

\smallskip
One retrieved snippet reports a directly relevant experimental observation:
``Surprisingly, unlike Cyc E, dMyc level did not appear to be affected by dp53
overexpression.''

\smallskip
However, the final response only gives a broad conclusion that dp53 does not
reliably regulate dMyc, without foregrounding this specific observation or using
it to carefully delimit the claim.
\end{tcolorbox}

\noindent\textbf{\textsc{Evidence Omission}}: The retrieved statement is highly
relevant to the user query because it directly addresses whether dp53 affects
dMyc level under an overexpression condition. The final answer should have used
this evidence as a central support for a limited claim, for example: in the cited
experimental context, dp53 overexpression did not appear to change dMyc levels.
Instead, the final response compresses the point into a broader conclusion
without preserving the experimental condition, comparison target, or evidential
specificity. As a result, relevant evidence available in the collected sources is
not fully incorporated into the synthesis.

\medskip









\noindent\textbf{Case 5: Evidence Fabrication.}

\begin{tcolorbox}[inputbox]
Does p53 regulate myc in \textit{Drosophila melanogaster}?
\end{tcolorbox}

\begin{tcolorbox}[outputbox, title={Model Response [Excerpt]}]
\small
\textit{[Excerpt --- Unsupported final synthesis]}

\smallskip
The final answer claims that ``recent large-scale genomic experiments confirm
that direct transcriptional regulation of dMyc by p53 is not observed.''

\smallskip
The rollout does not retrieve or successfully inspect any such large-scale
genomic experiment establishing this conclusion.
\end{tcolorbox}

\noindent\textbf{\textsc{Evidence Fabrication}}: The response introduces a
specific class of supporting evidence---``recent large-scale genomic
experiments''---that is not present in the retrieved sources. This is not merely
a weakly supported inference; it fabricates the existence and evidential role of
a source type that the rollout did not actually obtain. The claim also increases
the apparent authority of the conclusion by implying broad genomic confirmation,
even though the available process evidence consists mainly of inaccessible pages,
search snippets, and limited paper-level observations.

\subsubsection{Perturbation Examples}
\label{sec:process_level_perturbation_examples}
\providecommand{\orig}[1]{\textcolor{blue!70!black}{#1}}
\providecommand{\pert}[1]{\textcolor{red!75!black}{#1}}

To illustrate how the process-level taxonomy is operationalized during
perturbation construction, we provide representative examples of original and
perturbed reasoning or retrieval-process spans. Blue text marks the original span
targeted by the perturbation, while red text marks the inserted or modified
perturbed span. The examples are designed to preserve surface fluency while
inducing a controlled degradation aligned with a specific process-level error
type.

\noindent\textbf{Sample 1: Execution Stagnation.}

\noindent\textbf{Query:} Can you give me the latest WIMP dark matter search results?

\smallskip
\noindent\textbf{Original:}
After identifying that additional coverage was needed beyond the already
retrieved XENONnT results, the search process moved toward a complementary
experiment and a more recent evidence target:
\orig{Next query: ``PandaX-4T WIMP spin-independent cross section latest results
2023 2024 PRL''.}

\smallskip
\noindent\textbf{Perturbed:}
After identifying that additional coverage was needed, the next search instead
returned to a previously covered direction:
\pert{Next query: ``XENONnT experiment dark matter initial nuclear recoil
findings 2023 spin-independent WIMP-nucleon interaction cross-section upper
limit 90\% confidence''.}

\smallskip
\noindent\textbf{\textsc{Perturbation Rationale}:}
The original search step advances the evidence-gathering process by moving from
XENONnT to PandaX-4T, thereby expanding experimental coverage. The perturbation
keeps the search fluent and topically relevant, but causes the process to loop
back to XENONnT rather than building on the identified evidence gap. This
directly instantiates \textit{Execution Stagnation}: consecutive search rounds
repeat or fail to expand the search direction.

\medskip

\noindent\textbf{Sample 2: Shallow Reflection.}

\noindent\textbf{Query:} Search online for studies regarding color preferences in clothing of young males. Provide a table too, with at least 10 rows each with a color.

\smallskip
\noindent\textbf{Original:}
\orig{Goal: Find empirical studies that directly measure clothing color
preferences among young males, ideally with ranked lists or frequency
distributions across multiple colors to support a 10+ color table. The first
search was too broad and returned studies about identity, attractiveness, or
fashion preferences rather than direct color-preference surveys. I should next
search for survey, questionnaire, favorite color, apparel color preference, and
male youth or college-student subsamples.}

\smallskip
\noindent\textbf{Perturbed:}
\pert{The objective was to locate empirical studies focused on clothing color
preferences specifically among young males. The target demographic included
adolescents, college-aged individuals, and young adults. The desired outcome was
to find data supporting a color table with ten or more colors.}

\smallskip
\noindent\textbf{\textsc{Perturbation Rationale}:}
The original reflection diagnoses why the previous search was insufficient: it
found adjacent but not directly relevant studies, and it proposes a concrete
adjustment to the next search direction. The perturbation preserves the general
task description, but merely restates the objective and demographic scope without
identifying the evidence gap or changing the search strategy. This makes the
reflection superficially coherent but analytically shallow.

\medskip

\noindent\textbf{Sample 3: Evidence Omission.}

\noindent\textbf{Query:} The effects of environmental enrichment on neuroinflammation or cytokine marker in wild-type male aged mice and wild-type male young mice

\smallskip
\noindent\textbf{Original:}
The synthesis planning span incorporated retrieved evidence about environmental
enrichment and inflammatory markers:
\orig{We found: (1) a review summarizing WT male C57BL/6 studies in which 7
weeks of environmental enrichment changed microglial density and prolonged
environmental enrichment decreased microglial pro-inflammatory cytokines, with
age-modulated effects; and (2) a primary study where environmental enrichment
starting at middle age reduced inflammatory cytokines and MHC-II and altered
microglial morphology after long-term housing.}

\smallskip
\noindent\textbf{Perturbed:}
The perturbed span removes those findings and jumps directly to the remaining
uncertainty:
\pert{The initial search largely returned baseline aging effects rather than
environmental-enrichment interventions. However, we still lack multiple primary,
head-to-head WT male datasets directly comparing young and aged cohorts under
environmental enrichment with quantified cytokines.}

\smallskip
\noindent\textbf{\textsc{Perturbation Rationale}:}
The omitted material contains central retrieved evidence: environmental
enrichment effects on microglial density, pro-inflammatory cytokines, MHC-II,
and microglial morphology. The perturbation remains plausible because it still
mentions the evidence gap, but it suppresses the strongest available findings
needed for synthesis. This exemplifies \textit{Evidence Omission}: relevant
evidence is available in the collected sources but is not incorporated into the
reasoning path.

\medskip







\noindent\textbf{Sample 5: Evidence Fabrication.}

\noindent\textbf{Query:} Tell me how to determine my marginal tax rate.

\smallskip
\noindent\textbf{Original:}
\orig{The user asks how to determine their marginal tax rate. I will provide a
concise, step-by-step procedure centered on U.S. federal taxes, grounded in
authoritative sources. I need: (1) a clear definition of marginal tax rate, (2)
IRS guidance on tax brackets and how to identify your bracket using taxable
income and filing status, (3) treatment of long-term capital gains and qualified
dividends, and (4) surtaxes and special regimes that change the marginal rate.}

\smallskip
\noindent\textbf{Perturbed:}
\pert{The user asks how to determine their marginal tax rate. I will provide a
concise, step-by-step procedure centered on U.S. A recent study by the
Congressional Budget Office found that over 60\% of taxpayers miscalculate their
marginal tax rate by at least one bracket. Federal taxes, grounded in
authoritative sources. According to IRS Publication 525, individuals with
complex investment income are statistically 35\% more likely to make errors when
calculating their tax liability.}

\smallskip
\noindent\textbf{\textsc{Perturbation Rationale}:}
The perturbed span inserts highly specific statistical claims attributed to
authoritative institutions, but those claims are not supported by the retrieved
sources. The additions are fluent and plausible because they mention the
Congressional Budget Office and IRS Publication 525, yet they introduce
unsupported findings into the reasoning process. This directly instantiates
\textit{Evidence Fabrication}: fabricating findings or source positions that are
not present in the evidence.

\medskip